%% file: main.tex
\documentclass[acmlarge,nonacm]{acmart}
\AtBeginDocument{\providecommand\BibTeX{{\normalfont B\kern-0.5em{\scshape i\kern-0.25em b}\kern-0.8em\TeX}}}

\usepackage[export]{adjustbox}
\usepackage{tabularx}
\usepackage{makecell}
\usepackage{bm}
\usepackage{amsmath,amsfonts}
\usepackage{graphicx}
\usepackage{xcolor}
\usepackage{multirow}
\usepackage{hyperref}
\usepackage{booktabs}
\usepackage[shortlabels]{enumitem}
\usepackage{subcaption}
\usepackage{soul}
\usepackage{calc}
\usepackage[utf8]{inputenc}
\usepackage{color, colortbl}
\usepackage{tikz}
\usetikzlibrary{arrows.meta, positioning, fit, backgrounds, calc}

\setcopyright{cc}
\setcctype{by}
\acmJournal{IMWUT}
\acmYear{2027}

\newcommand{\projectname}{IMUdrape}

\newcommand{\best}[1]{\colorbox[HTML]{caebc0}{$#1$}}

\definecolor{ao(english)}{rgb}{0.0, 0.5, 0.0}

\usepackage{atbegshi}
\AtBeginShipout{\AtBeginShipoutAddToBox{%
  \begin{tikzpicture}[remember picture, overlay, red]
    \node[anchor=south, font=\large] at ([yshift=15mm]current page.south) {This manuscript is under review. Please write to zleng7@gatech.edu for up-to-date information};
  \end{tikzpicture}%
}}

\begin{document}

\title[IMUdrape: Physics-Based Garment-Mounted Virtual IMU Data]
{IMUdrape: Generating Garment-Mounted Virtual IMU Data from Text through Physics-Based Cloth Simulation}

\author{Zikang Leng}
\email{zleng7@gatech.edu}
\orcid{0000-0001-6789-4780}
\affiliation{\institution{School of Interactive Computing, Georgia Institute of Technology}
  \city{Atlanta, Georgia}
  \country{USA}}

\author{Thomas Plötz}
\email{thomas.ploetz@gatech.edu}
\orcid{0000-0002-1243-7563}
\affiliation{\institution{School of Interactive Computing, Georgia Institute of Technology}
  \city{Atlanta, Georgia}
  \country{USA}}

\renewcommand{\shortauthors}{Leng \& Plötz}

\begin{abstract}
\input{abstract/main}
\end{abstract}



\maketitle

\input{sections/intro/main}
\input{sections/related/main}
\input{sections/system/main}
\input{sections/evaluation/main}

\input{sections/discussion/main}
\input{sections/conclusion/main}
\begin{acks}
This work was partially supported by the NSF Research Fellowship under Grant No. DGE-2039655.
Any opinion, findings, and conclusions or recommendations expressed in this material are those of the author(s) and do not necessarily reflect the views of the National Science Foundation.
\end{acks}

\bibliographystyle{ACM-Reference-Format}
\bibliography{./bibs/ref}

\newpage
\appendix
\input{sections/appendix/main}

\end{document}

%% file: abstract/main.tex
\noindent
Wearable computing was envisioned as computation woven into what people literally wear, and inertial sensors are now moving into clothing itself, from sensorized sportswear to garments instrumented with hundreds of IMUs.
Sewn into clothing, a sensor can be comfortable and unobtrusive enough for continuous use, but it no longer moves rigidly with the body.
Loose fabric slides, twists and swings, so a garment-mounted sensor records the motion of the fabric as well as that of the body, and this extra motion differs from one garment to the next.
A model for a garment-mounted sensor therefore needs training data recorded with the garment it is deployed on, yet such labeled data are scarce, as they are for sensor-based activity recognition in general.
For body-attached sensors, virtual IMU data generated from videos or text have eased this scarcity.
Existing generators, however, attach their virtual sensors rigidly to the body, so they cannot reproduce how clothing moves.
We present \projectname{}, which closes this gap by generating garment-mounted virtual IMU data from text through physics-based clothes simulation.
\projectname{} synthesizes human motion from activity descriptions with a text-to-motion model, reconstructs the target deployment's garment as a sewing pattern from photographs and dresses the body in it, places virtual IMUs at the deployment's real sensor positions on the garment, and simulates the garment with an intersection-free contact solver, so that every simulation yields paired garment and body signals for the same motion.
On four real datasets with clothing-mounted sensors, \projectname{} improves on existing virtual-data approaches for both activity recognition and full-body pose estimation.
With one real subject, its data give a relative improvement in recognition macro-F1 of 7\% to 46\% over real data alone and of 2\% to 11\% over body-mounted virtual data from the same motion, and they remain ahead once all real subjects are used. When the garment is simulated on recorded motion, the pose error from loose-garment IMUs without any real training data falls by 30\% to 33\% relative to generated motion, and garment data remain 12\% more accurate than body data.
\projectname{} thus reduces the amount of real data that garment-mounted sensing requires, and it lays the groundwork for researchers who want to study smart-clothing sensing but have no recorded data of their own.

%% file: sections/intro/main.tex
\section{Introduction}
\label{sec:intro}

Body-worn inertial measurement units (IMUs) are one of the central sensing modalities of mobile and ubiquitous computing, supporting human activity recognition (HAR), full-body pose estimation and health monitoring~\cite{har_survey}.
Increasingly, these sensors are built into clothing, from commercial smart shirts and motion-capture suits~\cite{hexoskin,rokoko_smartsuit,xsens_link} to research garments that carry one to hundreds of IMUs~\cite{esfahani2019garments,lin2023loosegarment,datta2026stitchsense,jin2025bodysuit,mig_dataset,garmocap_gid}, because clothing is comfortable, unobtrusive and suited to long-term wear.
Most inertial-sensing methods, however, from sparse-IMU pose estimators~\cite{huang2018dip,yi2021transpose,yi2022pip} to generators of virtual IMU data~\cite{imusim,kwon2020imutube,leng2023imugpt}, assume that each sensor is fixed tightly to the body segment it measures, typically with straps~\cite{lip_cvpr2024}.
A garment-mounted sensor instead moves with the fabric, which slides, rotates, folds and swings, so its signal depends on the garment's cut, the fabric, the sensor's position and the activity~\cite{lorenz_loose_tight,jayasinghe2020clothing,lip_cvpr2024}.

This garment dependence sharpens the problem that limits all of sensor-based HAR, the lack of large, labeled datasets.
Data recorded with one garment may not generalize to another, even at the same body position and for the same activity~\cite{jayasinghe2020clothing,lorenz_loose_tight}, so a new garment may need new recordings.
For body-worn sensors, virtual IMU data derived from videos~\cite{kwon2020imutube,rey2019letthere,zhang2020deepgenerative,santhalingam2023asl}, motion capture~\cite{huang2018dip,uhlenberg2022surface,zhang2024unimts} or text~\cite{leng2023imugpt,leng2024benefit,haeusler2025text2imu} have eased this scarcity.
For garment-mounted sensors, existing approaches fall short.
Virtual IMU generators place their sensors rigidly on the body, so their data describe a body-mounted sensor rather than the garment-mounted sensor that is deployed.
Denoisers that map garment signals to body signals are learned from \emph{paired} garment and body recordings~\cite{garmocap_gid,lorenz_loose_tight}, which require laborious data collection for every garment.
Learned garment models~\cite{tailornet,lip_cvpr2024} need no recordings, but they predict the garment's shape from the pose without simulating its dynamics and cannot be matched to an arbitrary target garment.

In this paper we present \projectname{}, a pipeline that generates \emph{garment-mounted} virtual IMU data from text through physical simulation of the garment.
\projectname{} generates human motion from a textual activity description, dresses an SMPL body, a standard parametric model of the human body~\cite{smpl}, in a garment reconstructed to match the clothing of a target deployment~\cite{chatgarment,garmentcode}, places virtual IMUs at that deployment's real sensor positions, and simulates the garment with an intersection-free contact solver~\cite{ipc,ppf_solver}.
The virtual IMU signals are computed from small rigid patches of the simulated fabric at each sensor position.
Because both the motion and the garment are generated, \projectname{} can produce data for a wide variety of motions and garments.
Every simulation also yields \emph{paired} data, IMU signals on the body and on the garment for the same motion, which separates the effect of the garment from that of the motion.

We evaluate \projectname{} on four real datasets with clothing-mounted sensors, for activity recognition~\cite{mig_dataset,lip_cvpr2024,fip_dataset,garmocap_gid} and pose estimation~\cite{lip_cvpr2024,fip_dataset,garmocap_gid}, with three recognition models and two pose networks.
We compare it with real data alone, with body-mounted virtual IMU data from IMUGPT~\cite{leng2023imugpt,leng2024benefit} and with garment data from TailorNet~\cite{tailornet}, a learned garment model without dynamics.
\projectname{} itself provides three sources, the body signal (\projectname{} Body), the garment signal (\projectname{} Garment) and clips drawn from either (\projectname{} Body + Garment).

\projectname{} data are effective in both tasks.
For activity recognition with one real subject, \projectname{} Body + Garment improves macro-F1 by a relative 7.2\% to 46.3\% over real data alone across the four datasets, and by 2.1\% to 44.9\% over the other virtual sources, IMUGPT, TailorNet and \projectname{} Body. Even with all real subjects, the best \projectname{} source remains 1.0\% to 5.5\% above real data alone.
For pose estimation without real data, \projectname{} data give the lowest joint angle error on all three datasets, 7.8\% to 11.5\% below IMUGPT and 3.2\% to 10.0\% below TailorNet. When the garment is simulated on GarMoCap's recorded instead of generated motion, this error falls by a further 29.7\% to 33.3\%, and \projectname{} Garment stays 12.1\% below \projectname{} Body.
Across both tasks, garment physics matters most when real data are scarce and when the simulated motion is realistic.

Our contributions are as follows:
\begin{itemize}
  \item We present \projectname{}, a pipeline that generates garment-mounted virtual IMU data from text through physics-based cloth simulation, together with garments and sensor maps matched to the clothing and sensor layouts of four public datasets.
  \item We conduct a large-scale evaluation of garment-mounted virtual IMU data for activity recognition and pose estimation, against body-mounted and learned-garment virtual data, and characterize when garment physics matters: with scarce real data and with realistic motion.
  \item We provide empirical insights for practitioners on how much and which real and virtual data to collect, where on a garment and with how many sensors physics matters most, and how realistic motion and a matched garment add to the benefit.
\end{itemize}

%% file: sections/related/main.tex
\section{Related Work}
\label{sec:related}

Sensor-based human activity recognition (HAR) with body-worn inertial sensors is one of the central pillars of ubiquitous computing~\cite{bulling2014tutorial,har_survey,haresamudram2025past}.
Despite advances in deep~\cite{hammerla2016deep,deepconvlstm,murahari2018attention,abedin2021attend,zhou2022tinyhar} and self-supervised learning~\cite{saeed2019multitask,tang2021selfhar,xu2021limubert,haresamudram2022assessing} and in domain adaptation~\cite{chang2020domain}, its most persistent obstacle remains the lack of labeled training data, because recording and annotating body-worn sensor data is expensive, time-consuming and raises privacy concerns~\cite{ploetz2021pitfalls,ploetz2023moredata}.
Generating virtual training data through cross-modality transfer~\cite{kamboj2024crossmodal} is a promising response, and the one we follow.
Table~\ref{tab:related} positions \projectname{} relative to the most closely related virtual-data approaches.

\input{tables/related}

\subsection{Cross-Modality Transfer for Human Activity Recognition}
\label{sec:related:virtual}
Cross-modality transfer approaches address data scarcity by converting data from a modality that is plentiful into training data for a sensing modality that is not, most often inertial sensors.
IMUTube~\cite{kwon2020imutube,imutube_followup} extracts three-dimensional human motion from online videos with pose estimation and computes virtual accelerometry at body locations, and Rey et al.~\cite{rey2019letthere} derive training data for wearable activity recognition from monocular videos in a similar way.
Follow-up work improved the quality and calibration of video-derived data, showed that large virtual corpora enable more complex recognition models~\cite{imutube_followup,kwon2021complex}, and found that their benefit shrinks for fine-grained activities~\cite{leng2023utility}.
Accelerometry has also been synthesized from video with generative models~\cite{zhang2020deepgenerative}, video-derived data have supported eating detection, exercise and sign-language recognition~\cite{jain2022eating,xia2022springjoint,santhalingam2023asl}, and the same principle has produced radar data~\cite{ahuja2021vid2doppler,bhalla2021imu2doppler} and pre-training data from motion-capture archives~\cite{mahmood2019amass,zhang2024unimts} and image corpora~\cite{yoon2025vision2motion}.

With the emergence of large language models and text-driven motion synthesis~\cite{humanml3d,t2mgpt,tevet2023mdm,jiang2023motiongpt,zhang2024motiondiffuse}, language has become a source modality as well.
IMUGPT~\cite{leng2023imugpt,leng2024benefit} uses a language model to write diverse descriptions of how an activity can be performed, converts them into motion sequences with a text-to-motion model~\cite{t2mgpt}, and derives virtual IMU data with a biomechanical sensor simulator~\cite{imusim}.
Text2IMU~\cite{haeusler2025text2imu} animates body-surface models of different morphologies with text-generated motion, and Wonderwall~\cite{miao2026wonderwall} pre-trains a foundation model on physics-based simulated motion before refining it on real data.
Language has also been fused with sensor data for zero-shot recognition~\cite{su2025imuzero}, and MOSAIC~\cite{mosaic2026} trains a sensor--language model on a synthetic IMU--text corpus generated in the same way as IMUGPT's.

Across these approaches, the motion source varies, but the virtual sensor is rigidly attached to the body, at a joint of the skeleton~\cite{imusim} or on the body surface~\cite{uhlenberg2022surface}, with WIMUSim~\cite{wimusim} adding variability of body shape, sensor placement and hardware (Table~\ref{tab:related}).
This assumption is reasonable for strapped sensors but not for sensors on clothing, whose signal depends on how the garment moves relative to the body.
Like IMUGPT, \projectname{} generates motion from text, but it senses that motion through a physically simulated garment matched to the target deployment.

\subsection{Garment-Mounted and Loose-Clothing Inertial Sensing}
\label{sec:related:garment}
Smart textiles integrate sensing into everyday clothing~\cite{stoppa2014textiles,cherenack2012textiles}, and garments with embedded inertial sensors have been used to recognize activities and postures and to capture whole-body motion~\cite{esfahani2019garments,lin2023loosegarment,jin2025bodysuit,mig_dataset,garmocap_gid,datta2026stitchsense}.
Where a sensor sits, and how far it shifts, matters even for strapped sensors. A long line of work has studied how the placement and displacement of on-body sensors affect activity recognition~\cite{kunze2008displacement,banos2014displacement}, typically with benchmark datasets recorded with strapped sensors~\cite{chavarriaga2013opportunity,reiss2012pamap2,sztyler2016realworld}.
Clothing adds a further displacement, one that changes with the motion itself.
Paired recordings of garment-attached and strapped sensors show that clothing reduces the correlation between garment and body signals and adds high-frequency content, with effects that depend on the garment and the body location~\cite{jayasinghe2020clothing,jayasinghe2023,lorenz_loose_tight,kcl_loose_cloth}.
This additional motion does not necessarily harm recognition. Sensors mounted on fabric distinguish similar movements better than rigidly attached ones and can raise classification accuracy~\cite{michael2017}, which a probabilistic model explains by the larger statistical distance between movements that fabric motion creates~\cite{shen2023loose}.
Earlier work also estimated the posture-recognition performance of sensing garments by modeling garment wrinkles geometrically~\cite{harms2010}, but such models address static postures and do not generate training data for dynamic activities.

The effect of clothing is most visible in sparse inertial pose estimation, which reconstructs full-body motion from a handful of IMUs~\cite{vonmarcard2017sip,huang2018dip,yi2021transpose,yi2022pip,jiang2022tip,mollyn2023imuposer}.
These methods assume that the IMUs are strapped tightly to the body, and they are trained largely on virtual IMU data synthesized from motion capture~\cite{mahmood2019amass}.
Loose Inertial Poser~\cite{lip_cvpr2024} and related work~\cite{fip_dataset,cip_dataset} show that their pose error rises markedly when the same IMUs are instead attached to loose garments, because the sensors then rotate and shift relative to the limbs they are meant to track.
To close this gap, LIP trains partly on loose-garment data synthesized with TailorNet~\cite{tailornet}.
TailorNet is a neural network that predicts how a garment of one of a few types drapes for the current body pose and shape. Trained on static poses, it models neither the garment's dynamics nor its contact with the body.
GaIP~\cite{gaip} generates loose and tight IMU pairs on motion-capture data, using TailorNet garments for the sensor orientation and a diffusion model for the loose signal.
Closest to ours, Haeusler et al.~\cite{haeusler2025garment} dress SMPL bodies of eight sizes in tight and loose TailorNet garments, a long-sleeve shirt and long pants, animate them with text-generated motion and place virtual IMUs on the body and on the garment.
They show that clothing changes the synthesized signals, most for loose garments and fast activities, and that a recognition model trained on garment data transfers better to the RealWorld dataset~\cite{sztyler2016realworld} than one trained on body data.
Because TailorNet predicts each frame from the pose, their garments have no inertia or lag, and the authors name physical modeling of gravity and inertia as future work. They also do not match the garment to the clothing of a target deployment, and they address activity recognition only.
A different approach removes the garment's effect instead of synthesizing it. GarMoCap~\cite{garmocap_gid} learns a garment-IMU denoiser (GID) that maps garment signals to body signals, from paired real recordings.
These approaches require real paired data or rely on learned garment shapes without dynamics, and none matches the garment of an arbitrary target deployment.
\projectname{} simulates the garment's dynamics from a sewing pattern, so it needs no real paired data and can represent virtually any everyday garment.

\subsection{Garment Modeling and Cloth Simulation}
\label{sec:related:cloth}
Simulating a specific garment first requires a model of its shape, and sewing patterns provide one. Like a tailor's pattern, they describe a garment as flat fabric panels and the seams that join them.
Parametric systems such as GarmentCode~\cite{garmentcode} describe garments by programmable patterns, and learned models reconstruct such patterns from three-dimensional scans~\cite{korosteleva2022neuraltailor} or, with vision--language models such as ChatGarment~\cite{chatgarment}, from photographs, which makes it possible to reconstruct a specific garment; Dress-1-to-3~\cite{li2025dress1to3} further refines such patterns with a differentiable simulator into simulation-ready garments from a single image.

Learned garment models predict the shape of clothing directly, from images~\cite{bhatnagar2019mgn}, as an offset from the body surface~\cite{ma2020cape} or for a given body pose~\cite{tailornet,santesteban2019learning}. They capture the shape a garment would settle into, but not how it lags, swings and slides.
Neural simulators learn these dynamics as well~\cite{santesteban2022snug,bertiche2022neuralclothsim,grigorev2023hood}, but all of these models are limited to the garments and motions they were trained on, which is why we turn to physics-based simulation.

Physics-based cloth simulation instead computes the motion of the fabric from its material properties, gravity and contact, and has a long history in computer graphics~\cite{baraff1998large,bridson2002robust,narain2012adaptive,macklin2016xpbd}.
Incremental Potential Contact (IPC)~\cite{ipc} treats contact as an energy that grows sharply as two surfaces approach, which guarantees that they never pass through each other.
Its extension to thin surfaces~\cite{li2021cipc} brings this guarantee to fabric, and GPU solvers of this family~\cite{ppf_solver} make it fast enough for large-scale use.
Closer to our setting, simulated loose garments have been used to study how clothing affects optical motion capture~\cite{drapemocap}.
\projectname{} combines sewing-pattern reconstruction with physics-based simulation. It reconstructs each target garment as a GarmentCode pattern~\cite{garmentcode} with ChatGarment~\cite{chatgarment} and simulates it with the ZOZO contact solver~\cite{ppf_solver}, a GPU solver of the IPC family~\cite{ipc,li2021cipc}, which keeps the garment intersection-free even in fast motion and in close contact with the body.

%% file: tables/related.tex
\begin{table}[!tbp]
  \caption{Overview of approaches that generate virtual inertial data, and how they model the attachment of the sensor. Sensor: where the virtual sensor sits; Body: rigidly attached to a body segment; Garment: on modeled clothing. Dynamics: the garment's motion relative to the body is simulated. Matched: the garment can be matched to the clothing of a target deployment.}
  \label{tab:related}
  \centering
  \footnotesize
  \begin{tabular}{@{}llllccl@{}}
    \toprule
    System & Motion source & Sensor & Garment model & Dynamics & Matched & Tasks \\
    \midrule
    IMUTube~\cite{kwon2020imutube,imutube_followup} & Video & Body & -- & -- & -- & HAR \\
    Rey et al.~\cite{rey2019letthere} & Video & Body & -- & -- & -- & HAR \\
    IMUGPT~\cite{leng2023imugpt,leng2024benefit} & Text & Body & -- & -- & -- & HAR \\
    UniMTS~\cite{zhang2024unimts} & Motion capture & Body & -- & -- & -- & HAR \\
    MOSAIC~\cite{mosaic2026} & Text & Body & -- & -- & -- & HAR \\
    LIP synthetic data~\cite{lip_cvpr2024} & Motion capture & Garment & Learned (TailorNet) & -- & -- & Pose \\
    GaIP~\cite{gaip} & Motion capture & Garment & \makecell[l]{Learned (TailorNet)\\+ diffusion} & -- & -- & Pose \\
    Haeusler et al.~\cite{haeusler2025garment} & Text & \makecell[l]{Body,\\garment} & Learned (TailorNet) & -- & -- & HAR \\
    \projectname{} (ours) & \makecell[l]{Text or\\recorded motion} & \makecell[l]{Body,\\garment} & \makecell[l]{Sewing pattern\\+ physics} & yes & yes & HAR, pose \\
    \bottomrule
  \end{tabular}
\end{table}

%% file: sections/system/main.tex
\section{\projectname{}: Generating Garment-Mounted Virtual IMU Data}
\label{sec:system}
\input{figure/teaser/teaser}

\projectname{} turns a textual description of an activity into paired body- and garment-mounted virtual IMU data.
Figure~\ref{fig:teaser} gives an overview of the pipeline.
Like IMUGPT~\cite{leng2023imugpt,leng2024benefit}, it starts from language.
A large language model writes varied descriptions of each target activity (for example, \emph{``a person jogs at an easy pace, arms swinging loosely''}), and a text-to-motion model turns each description into a 3D motion of the body.
We use HY-Motion~1.0~\cite{hymotion}, a flow-matching text-to-motion model whose output we map to SMPL~\cite{smpl}, a parametric model of the human body surface. 
Everything after the body motion is new in \projectname{} and is what body-mounted virtual IMU generators lack.
\emph{Garment generation} (Section~\ref{sec:system:garment}) builds a simulation-ready garment from a photograph or a description of the target garment.
The garment consists of a sewing pattern, whose flat panels are stitched together around the body into a 3D garment, and the mechanical properties of its fabric.
\emph{Garment simulation} (Section~\ref{sec:system:sim}) moves the garment physically with the generated body, so that it slides, twists and swings relative to the body as a real garment does.
\emph{Garment IMU} (Section~\ref{sec:system:sensing}) reads virtual sensors off the simulated garment at any point of the garment, including the sensor positions of a real smart garment, and pairs each with a sensor on the body beneath it.
The pipeline runs without manual intervention. Further details of each step and all parameter values are given in Appendix~\ref{app:sim}.

\subsection{Garment Generation}
\label{sec:system:garment}
Unlike in body-mounted virtual IMU data, the garment is a free variable of \projectname{}.
The same motion can be simulated with any garment, so the virtual wearer can be dressed in a garment that matches the deployment.
To simulate such a garment, we use a cloth simulator.
It needs three inputs: (i) the garment's sewing pattern, which fixes its cut and fit, (ii) the 3D garment on the body, stitched together from the pattern, and (iii) the mechanical properties of its fabric.
The following subsections describe how \projectname{} produces each of them.

\input{figure/garmentgen/garmentgen}

\subsubsection{Sewing-Pattern Estimation}
We estimate the sewing pattern with ChatGarment~\cite{chatgarment}, a vision--language model that reads a photograph of a garment, a text description, or both, and outputs the design of the garment.
The design has two parts.
Its discrete choices, such as the garment type, the collar, the cuffs and whether the front is open, are written as text.
Its 76 continuous values $\mathbf n$, such as lengths, widths and angles, are regressed from the model's internal state by a small network, because language models are poor at producing precise numbers (Appendix~\ref{app:pipeline}).
For a photograph, the model first describes the garment in words, part by part, and then outputs the design given both the photograph and this description.

\subsubsection{Sewing-Pattern Program}
The design $\theta$ (Figure~\ref{fig:garmentgen}a--c), which consists of the categorical choices and the continuous values $\mathbf n$, is the input of a GarmentCode program~\cite{garmentcode}.
GarmentCode is a programming language designed specifically for sewing patterns.
It builds a garment from components, such as a torso, two sleeves and a collar, each with its own parameters, and builds components from panels and panels from edges.
The program $\mathcal G$ maps the design and a vector $\mathbf b$ of body measurements (girths, lengths and heights) to a sewing pattern, a set of panels and stitches,
\begin{equation}
  \mathcal P = \mathcal G(\theta, \mathbf b) = \big(\{(\Omega_k, R_k, \mathbf t_k)\}_{k}, \{s_l\}_{l}\big).
\end{equation}
Each panel $\Omega_k\subset\mathbb R^2$ is a closed outline of straight edges and Bézier curves in the plane of the fabric.
The rotation $R_k$ and translation $\mathbf t_k$ place the panel in 3D around the body, so that a point $(u,v)\in\Omega_k$ starts at
\begin{equation}
  \mathbf x = R_k\,(u, v, 0)^\top + \mathbf t_k .
\end{equation}
Each stitch $s_l$ joins an edge of one panel to an edge of another (or the same) panel.
Because the measurements $\mathbf b$ are computed from the SMPL body mesh, the same design is automatically sized to the simulated body (Figure~\ref{fig:garmentgen}d).

\subsubsection{Virtual Sewing}
Each panel is triangulated as a flat mesh.
This flat 2D geometry is the \emph{rest shape} of the fabric, the shape in which it is unstretched, and the simulator measures stretch and bending relative to it.
Each stitch is turned into many small links that join every vertex on one edge to the matching point on the other edge.
We close these links in the same cloth simulator that later moves the garment (Section~\ref{sec:system:sim}), with the links acting as zero-length springs (Appendix~\ref{app:pipeline}).
The body stays static in the dressing pose, and inertia and gravity are switched off, so that the panels are pulled shut slowly and evenly.
Sewing succeeds only if the panels start outside the body and on both sides of the limb they wrap, and \projectname{} places them this way automatically.
For the same reason, inner garments are sewn before outer ones, so that each garment starts outside those beneath it.
The jacket, for example, is sewn over the already-sewn trousers, which act as an obstacle.

The fabric parameters (density, bending stiffness and damping) are set from the fabric type of the garment, using measured properties of typical fabrics of that type (Appendix~\ref{app:sim}). With these parameters, the sewn garment then settles under gravity with friction against the body.
The settled garment is the start state of every simulation with that garment (Figure~\ref{fig:garmentgen}h).
The garments we generated for the datasets of this paper are shown in Appendix~\ref{app:garments}.

\subsection{Garment Simulation}
\label{sec:system:sim}
The garment is simulated on the moving body with a contact solver that never lets garment and body interpenetrate.
Such a solver needs a physically clean body, and generated motion does not always provide one.
We therefore first describe how \projectname{} prepares the body (Figure~\ref{fig:garmentsim}) and then the solver.

\input{figure/garmentsim/garmentsim}

\subsubsection{Kinematic Body and Collider}
Two defects of generated motion have to be handled before the garment can be simulated.
First, body parts pass through each other.
With the default SMPL body shape, the upper arm of a generated walk sits inside the torso in most frames, because the motion model was trained on a slimmer body than SMPL (Appendix~\ref{app:collider}).
Fabric caught between two overlapping body parts has no valid position.
Second, linear blend skinning, the standard way of deforming the body mesh with the skeleton, folds the body surface at a flexed elbow, hip or crotch into a sharp, hook-shaped cusp.
Garment vertices get trapped in this cusp.

\projectname{} therefore uses two copies of the body.
The \emph{kinematic body} follows the generated motion exactly and carries the body sensors.
The \emph{collider} is the surface that the garment touches.
It starts as a copy of the kinematic body, and wherever the motion is not physically clean, it is deformed in that frame, much as soft tissue would be (Figure~\ref{fig:garmentsim}a,b).
Because the body sensors stay on the kinematic body, these deformations do not change the body IMU.

The collider removes the first defect by pushing arm and torso apart.
Wherever an arm vertex is closer to the torso than a target clearance, the missing clearance is split between arm and torso, which move apart along the line between them (Appendix~\ref{app:pipeline}).
These displacements are smoothed over the surrounding surface, so that the collider gets a gentle dent that moves with the limb, and the procedure is repeated until arm and torso are clear in every frame.

The collider removes the second defect by filling the joint creases. Imagine rolling a ball of radius $r$ over the outside of the two limbs that meet at a flexed joint.
The ball cannot enter a crease narrower than itself.
The surface it touches is therefore the limbs, with the crease filled by a smooth round of radius $r$, and this filled shape becomes the collider at the joint.
In practice, every collider vertex at the bottom of the crease, where the ball cannot reach, is moved outward onto the round.
In image processing, this operation is called a morphological closing.
As the joint flexes, the round grows smoothly and pushes the fabric out of the crease instead of pinching it.
Finally, because the garment is sewn in a fixed dressing pose, each motion is preceded by a short transition from that pose into the motion's first frame.

\subsubsection{Contact-Resolving Garment Dynamics}
We simulate the garment with the ZOZO contact solver~\cite{ppf_solver}, a GPU solver from the family of Incremental Potential Contact (IPC)~\cite{ipc,li2021cipc}.
These solvers advance the simulation by one time step by minimizing an energy.
For vertex positions $\mathbf x$ and time step $\Delta t$, each step finds the minimizer of
\begin{equation}
  \psi_{\mathrm{total}}(\mathbf x) = \psi_{\mathrm{dyn}}(\Delta t, \mathbf x) + \sum_i \psi_{\mathrm{weak}}\big(g_i, \hat g, \bar\kappa_i\big).
\end{equation}
The first term moves the fabric as momentum and its own elasticity dictate,
\begin{equation}
  \psi_{\mathrm{dyn}}(\Delta t, \mathbf x) = \frac{1}{2\Delta t^2}\,\|\mathbf x - \tilde{\mathbf x}\|^2_{M} + \Psi_{\mathrm{el}}(\mathbf x).
\end{equation}
Here $\tilde{\mathbf x}$ is where each vertex would end up if it kept its current velocity under gravity alone~\cite{ipc}, and the first part penalizes leaving that path, weighted by the vertex masses $M$.
The elastic energy $\Psi_{\mathrm{el}}$ penalizes stretching, shearing~\cite{baraff1998large} and bending the fabric.
The minimum is where the fabric keeps as much of its momentum as its elasticity allows.

The second term keeps garment and body apart.
Every contact $i$, between garment and body or between two parts of the garment, adds a barrier on its gap $g_i$ that grows as the gap closes.
IPC uses a logarithmic barrier, which becomes infinitely stiff as the gap closes and slows the solver down.
The ZOZO solver uses a cubic barrier instead~\cite{ppf_solver},
\begin{equation}
  \psi_{\mathrm{weak}}(g_i, \hat g, \bar\kappa_i) =
  \begin{cases}
    -\dfrac{2\bar\kappa_i}{3\hat g}\,(g_i - \hat g)^3, & g_i \le \hat g,\\[4pt]
    0, & \text{otherwise},
  \end{cases}
  \label{eq:barrier}
\end{equation}
which is zero beyond the activation gap $\hat g$ and rises smoothly below it (Figure~\ref{fig:garmentsim}c).
The solver sets the stiffness $\bar\kappa_i$ of each barrier automatically from the mass and the stiffness of the fabric at the contact, so that the barrier is strong enough to stop the fabric's momentum and to resist the fabric being pressed onto the body~\cite{ppf_solver} (Appendix~\ref{app:sim}).
No manual tuning is needed.
Each step is solved by Newton's method.
Before accepting a step, the solver checks along the whole path of every vertex that no part of the garment passes through the body or through the garment itself, so every accepted step is free of intersections.
Beyond contact, three settings shape how the garment moves on the body. Friction between garment and body follows the lagged dissipative model of IPC~\cite{ipc}, stiffness-proportional damping suppresses unphysical high-frequency vibration of the fabric, and a soft pin holds the trousers at the waist, as a belt would.

\subsection{Garment IMU}
\label{sec:system:sensing}
\input{figure/closeup/closeup}
A real garment IMU sits on a small rigid circuit board sewn onto the fabric.
We therefore model each virtual garment sensor as a small rigid patch of fabric (Figure~\ref{fig:closeup}).
To reproduce the sensor layout of a real smart garment, each virtual sensor is placed at the point of the garment closest to the body vertex under the corresponding real sensor.
This point is chosen in the settled start state, as if the sensor had been sewn over that spot while the wearer stood.

The fabric under a sensor deforms in every frame, so its vertices do not share one rotation and position.
Like a rigid board sewn onto the fabric, the virtual sensor follows the rigid motion that best matches the fabric beneath it.
Let $\mathcal P$ be the garment vertices within radius $r$ of the sensor point, $\mathbf x^0_j$ their positions in the start state, $\mathbf x_j(t)$ their positions at time $t$ and $a_j$ their areas.
The sensor's rotation $R(t)$ and position $\mathbf p(t)$ minimize the area-weighted distance between the moved patch and the fabric,
\begin{equation}
  \big(R(t), \mathbf p(t)\big) = \arg\min_{R\in SO(3),\,\mathbf p}\; \sum_{j\in\mathcal P} a_j\,\big\| R\,(\mathbf x^0_j - \bar{\mathbf x}^0) + \mathbf p - \mathbf x_j(t) \big\|^2 ,
\end{equation}
with $\bar{\mathbf x}^0$ the patch centroid in the start state.
This is the Kabsch problem~\cite{kabsch1976}, which has a closed-form solution from the singular value decomposition of the weighted cross-covariance of the patch.
The radius $r$ sets how much the patch smooths the fabric motion. A single vertex would overestimate how much the fabric rotates locally, and a larger patch over-smooths it. We use $r=2.5$\,cm, about the size of a small IMU board.
The sensor axes are fixed to the fabric (Figure~\ref{fig:closeup}).
The $z$ axis is the outward normal of the fabric, and the $x$ axis points along the vertical direction of the flat panel, so that a sensor sewn upright on the garment stays aligned with the fabric as it moves.
With $S^0$ being the sensor axes in the start state, $S(t)=R(t)\,S^0$ is the sensor frame at time $t$.
With the sensor's position and frame known over time, the accelerometer measures the sensor's acceleration minus gravity, $\mathbf a(t)$, and the gyroscope its angular velocity, $\boldsymbol\omega(t)$,
\begin{equation}
  \mathbf a(t) = S(t)^\top\big(\ddot{\mathbf p}(t) - \mathbf g\big), \qquad
  [\boldsymbol\omega(t)]_\times = S(t)^\top \dot S(t),
\end{equation}
with derivatives computed from the 120\,Hz simulation output and $\mathbf g$ the gravity vector.
The paired \emph{body} sensor is a rigid patch of the same radius on the kinematic body at the same body vertex, and its accelerometer and gyroscope signals are computed in the same way.
Every simulation therefore yields garment and body IMU signals for exactly the same motion (Appendix~\ref{app:sim}).

Pose estimation also needs each sensor's orientation relative to the bone beneath it, such as the forearm for a sensor on the sleeve cuff.
Pose datasets obtain it by calibration. The wearer holds a known pose at the start of a recording, and the rotation between each sensor and its bone in that pose is stored and treated as fixed.
For a garment sensor, this rotation depends on how the garment hangs during the pose, so \projectname{} reproduces the calibration with a simulated T-pose hold before each motion (Appendix~\ref{app:pipeline}).

Finally, the signals so far are those of an ideal sensor, and real IMUs have noise and other errors.
\projectname{} adds the standard error model of a MEMS IMU~\cite{woodman2007inertial,elsheimy2008allan},
\begin{equation}
  \tilde{\mathbf a}(t) = (I + \mathrm{diag}(\mathbf s))\,M_a\,\mathbf a(t) + \mathbf b + \boldsymbol\eta(t),
\end{equation}
with scale errors $\mathbf s$, axis misalignment $M_a$, bias $\mathbf b$ and white noise $\boldsymbol\eta$, and likewise for the gyroscope.
The error magnitudes are fitted to still windows of real recordings.
Because a real sensor is never sewn on exactly as designed, \projectname{} also rotates each garment sensor by a small random mounting rotation.

%% file: figure/teaser/teaser.tex
\begin{figure*}[!tbp]
\centering
\includegraphics{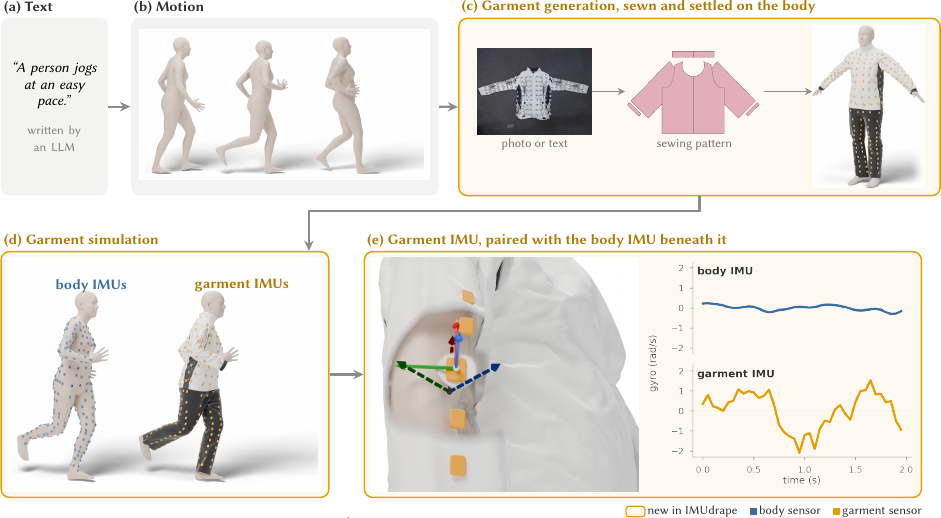}
\caption{\projectname{} generates \emph{garment-mounted} virtual IMU data from text. A text-to-motion model turns an activity description into body motion (a, b), as in body-mounted generators. \projectname{} then generates a garment from a photo or a description of the target garment (c), simulates it on the moving body (d), and reads virtual IMUs off the simulated garment at any point of the garment, each paired with an IMU on the body beneath it (e). Orange panels mark the parts that are new in \projectname{}, and blue and orange squares mark body and garment sensors. The garment sensor slides and twists relative to the body. Within 1.6\,s, its frame (solid axes) turns from 11$^\circ$ to 42$^\circ$ away from the frame of the body sensor beneath it (dashed axes), so the paired signals of the same motion differ strongly.}
\label{fig:teaser}
\end{figure*}

%% file: figure/garmentgen/garmentgen.tex
\begin{figure*}[!tbp]
\centering
\includegraphics{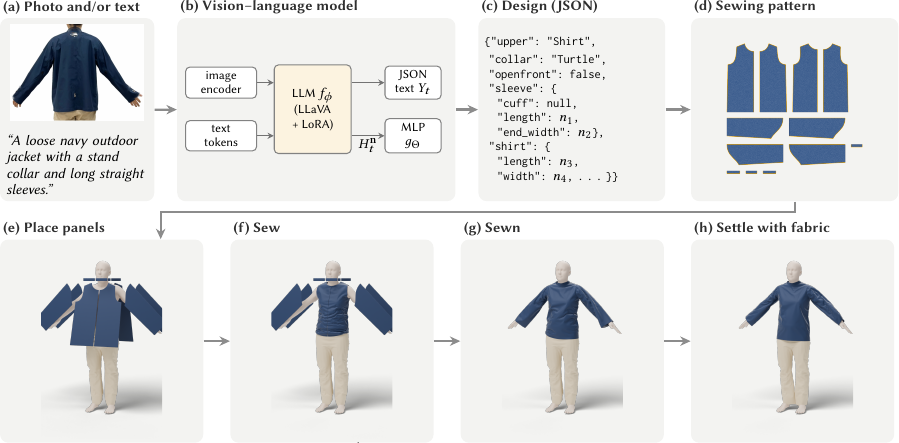}
\caption{Garment generation. (a--c) A vision--language model~\cite{chatgarment} turns a photograph and/or a description of the garment into a design (photograph adapted from LIP~\cite{lip_cvpr2024}, annotations removed). The design consists of categorical choices (garment type, collar, cuffs) as JSON text and 76 continuous values (lengths, widths, angles) regressed from the hidden state of an end token. (d) A GarmentCode program~\cite{garmentcode} turns the design and the body measurements into 2D sewing panels. Orange edges are sewn to other panels. (e--h) The panels are placed around the body, sewn shut in the cloth simulator without inertia or gravity, and settled under gravity with the fabric's properties. The settled garment is the start state of every simulation. The garment shown is the LIP jacket, sewn over trousers.}
\label{fig:garmentgen}
\end{figure*}

%% file: figure/garmentsim/garmentsim.tex
\begin{figure*}[!tbp]
\centering
\includegraphics{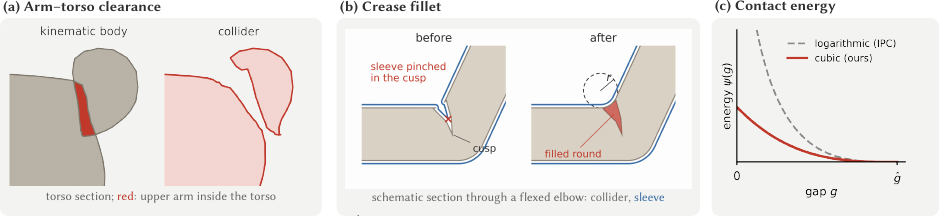}
\caption{Garment simulation. (a) In a generated motion (drumming), the upper arm of the kinematic body, which carries the body IMU, lies inside the torso (red, left). In the collider, the surface that the garment touches, arm and torso are pushed apart (right). (b) Schematic: skinning folds the body surface at a flexed elbow into a narrow cusp, and the sleeve is pinched in it, where the solver cannot resolve the contact (left). A ball of radius $r$ rolled over the limbs cannot enter the crease; the region it cannot reach (red) is added to the collider, and the sleeve runs over the smooth round (right). (c) Contact is enforced by a barrier that is zero beyond the gap $\hat g$. The cubic barrier~\cite{ppf_solver} stays finite as the gap closes, unlike the logarithmic barrier of IPC~\cite{ipc}.}
\label{fig:garmentsim}
\end{figure*}

%% file: figure/closeup/closeup.tex
\begin{figure}[h]
  \centering
  \includegraphics[width=\linewidth]{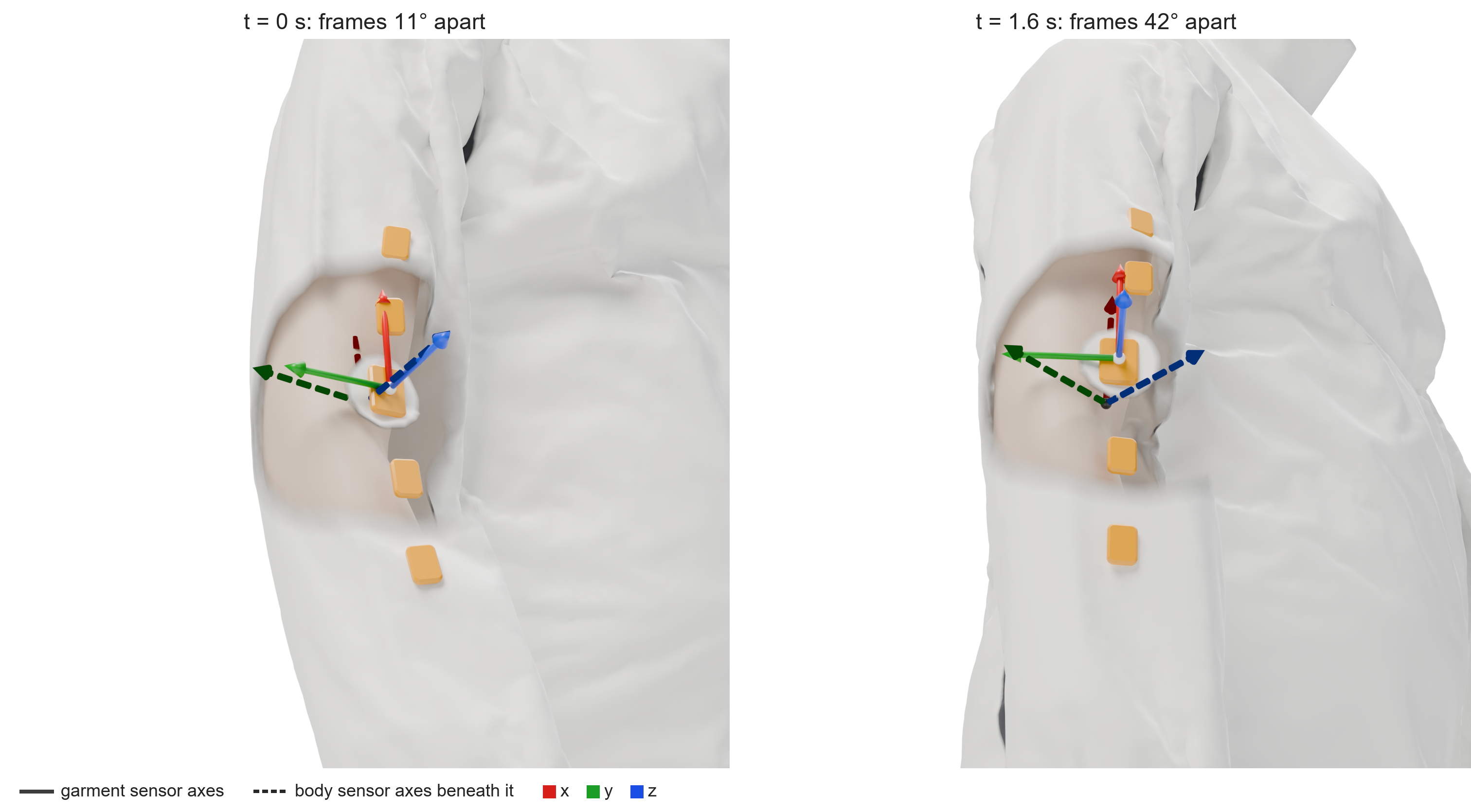}
  \caption{A garment sensor on the upper sleeve (solid axes) and the body sensor beneath it (dashed axes), seen through a cut-away in a generated folding motion. The garment sensor's $z$ axis is the outward normal of the fabric. At the start of the motion, the two frames are 11$^\circ$ apart (left). As the sleeve slides and twists, the garment frame rotates 42$^\circ$ away from the body frame within 1.6\,s (right).}
  \label{fig:closeup}
\end{figure}

%% file: sections/evaluation/main.tex
\section{On the Effectiveness of Garment-Mounted Virtual IMU Data}
\label{sec:eval}

A sensor sewn into a garment moves with the fabric, so a model for it needs training data recorded with that garment, and such data are scarce.
In this section, we test whether models trained with \projectname{}'s virtual data perform well on the IMU data of real garments.
We consider a deployment in which a garment with built-in IMUs should recognize its wearer's activities or estimate the wearer's full-body pose, for example to follow rehabilitation exercises at home or an athlete's technique without a motion-capture suit (Section~\ref{sec:eval:setup}), but labeled recordings exist for only a few people, or none.
\projectname{} generates virtual IMU data for the activities of interest, on a simulated copy of the garment and at the positions of its sensors.
We train a model on the virtual data together with whatever real data are available, and test it on real recordings of a person who was not part of the training data and who wears the real sensor garment.
We compare it with a model trained on real data alone and with models trained on virtual data from existing generators.
Section~\ref{sec:eval:setup} describes the datasets, the virtual data and the protocol that both tasks share, and Sections~\ref{sec:eval:har} and~\ref{sec:eval:pose} report the results of each task.

\subsection{Experimental Setup}
\label{sec:eval:setup}

\subsubsection{Datasets}
We evaluate on four public datasets recorded with inertial sensors on everyday garments.
The MIG dataset~\cite{mig_dataset} was recorded with a jacket and trousers carrying 396 sewn-in IMUs.
The LIP~\cite{lip_cvpr2024}, FIP~\cite{fip_dataset} and GarMoCap~\cite{garmocap_gid} datasets were recorded with four to six IMUs on loose jackets, and in GarMoCap also on loose trousers, together with the wearer's pose.
The activities range from locomotion (walking, running, stairs) and sports (boxing, ping-pong, basketball, jumping rope) to everyday tasks (eating, typing, ironing, vacuuming). MIG has 14 activity classes, LIP 5, and FIP and GarMoCap 10 each.
Table~\ref{tab:datasets} in Appendix~\ref{app:datasets} lists the subjects, sensors, activities and recorded minutes of every dataset, and the amount of virtual data we generate for each.

All four datasets label the activities, so we use all of them for activity recognition, the established benchmark for virtual IMU data~\cite{kwon2020imutube,leng2023imugpt}.
In addition to activity recognition, we consider full-body pose estimation on the LIP, FIP and GarMoCap datasets, which also record the wearer's pose.
Full-body pose is the most detailed description of how a person moves. It is the basis of motion capture for sports and rehabilitation, of avatars in virtual and augmented reality, and of finer-grained activity and health analyses than a fixed set of activity labels allows~\cite{huang2018dip,yi2021transpose}.
Sensors built into everyday clothing would make such motion capture available outside the lab, without straps or a suit~\cite{lip_cvpr2024,garmocap_gid}.
Pose labels, however, require an optical or inertial motion-capture system during recording, so labeled garment data for pose estimation are even scarcer than for activity recognition.
This makes pose estimation a demanding test of whether virtual garment data can stand in for real recordings.

\subsubsection{Virtual Data and Protocol}
\label{sec:eval:main}
\label{sec:eval:baselines}
For every dataset, we generate 20 text descriptions of each of its activities, generate the motion and simulate it on the garment matched to the dataset's clothing (Table~\ref{tab:garments} in Appendix~\ref{app:garments}).
This gives 89 minutes of virtual data for MIG, 12 for LIP, 24 for FIP and 21 for GarMoCap, 146 minutes in total (Table~\ref{tab:datasets} in Appendix~\ref{app:datasets}).
The virtual data are thus smaller than the real recordings of each dataset, between about a quarter and a tenth of their size.

Existing virtual IMU data place the sensors on the body, and the only established way to obtain garment motion without physics simulation is a learned garment model such as TailorNet~\cite{tailornet}.
We therefore compare \projectname{} with one baseline for each approach. \projectname{} itself provides three sources from the same simulations:
\begin{itemize}
  \item \textbf{IMUGPT}~\cite{leng2023imugpt,leng2024benefit} generates motion from the same descriptions with its motion synthesis model T2M-GPT~\cite{t2mgpt} and computes body IMU at the skeleton joints with IMUSim~\cite{imusim}. As calibration, IMUGPT fits the signal distribution of each activity class to the real training data; without real training data, its data are used uncalibrated.
  \item \textbf{TailorNet}~\cite{tailornet} is a network that predicts the garment mesh from the body's pose and shape. It was trained on physics simulations of garments on single poses and predicts each frame independently, so the garment follows the body without inertia or delay. LIP's synthetic loose-garment data are built on it~\cite{lip_cvpr2024}. We run it on our motion and read the sensors from the predicted garment as in \projectname{}. 
  \item \textbf{\projectname{} Body} uses the signal of a sensor on the body at the same position, which every \projectname{} simulation also gives. It separates the effect of the garment from that of the motion. A body-worn sensor is strapped in place, so it receives the sensor errors but not the mounting rotation of a garment sensor.
  \item \textbf{\projectname{} Garment} simulates the garment on our motion and reads the sensors from the simulated fabric.
  \item \textbf{\projectname{} Body + Garment} pools the body and garment data of the same clips and subsamples the pool to the size of the other sources, so that each motion appears once, as either its body or its garment signal.
\end{itemize}
Every source has the same number of clips per activity, where a clip is one generated motion sequence with its virtual IMU signals, so all sources have the same amount of virtual data.
Every dataset uses leave-one-subject-out cross-validation over all of its subjects.
We train with real data from 0 (virtual data only), 1, 2 or all training subjects.
No real subjects corresponds to zero-shot and one or two subjects to few-shot learning, the low-data regimes in which a new garment is deployed. All subjects show whether virtual data still help once real data are plentiful.
For each model, we repeat the cross-validation with three random seeds and report the average and the standard deviation across the three runs.

\subsection{Activity Recognition}
\label{sec:eval:har}

\subsubsection{Experimental Setting}
We classify 2\,s windows of garment IMU signal into the activities of each dataset.
The LIP, FIP and GarMoCap datasets provide accelerometer and orientation data without a gyroscope, so on these datasets we use the accelerometer signals for recognition.
On the MIG dataset, we use six of the 396 sensors, on the right side where the fabric hangs free: the back of the forearm sleeve, the side panel below the armpit, the front hip below the belt, the buttock and the front and back of the ankle.
We use three recognition models, DeepConvLSTM~\cite{deepconvlstm}, DeepConvLSTM with attention~\cite{murahari2018attention} and TinyHAR~\cite{zhou2022tinyhar}. Appendix~\ref{app:models} describes the models, and Appendix~\ref{app:sites} the sensor mapping of each dataset.
The model is trained on the real and virtual windows together, as in earlier work on virtual IMU data for activity recognition~\cite{kwon2020imutube,leng2023imugpt}.

Sensor signals differ between people and with how each person wears the garment, so we use test-time adaptation~\cite{li2018adabn,wang2021tent} by per-subject standardization.
Every subject's channels are standardized with its own mean and standard deviation, and the virtual data with theirs.
For the test subject, these statistics come from the unlabeled earlier half of each recording, and the later half is used for evaluation.
This mirrors a real deployment, in which a system can collect unlabeled data from a new user before it has to recognize that user's activities, as in unsupervised adaptation of activity recognition to new users~\cite{chang2020domain}.
We report the macro-F1 score.

\subsubsection{Results}
Table~\ref{tab:har_main} shows the results.

\input{tables/har_main}

An \projectname{} source gives the best result in every setting, and the gain is largest when real data are scarce.
With one real subject, \projectname{} Body + Garment improves macro-F1 over real data alone by a relative 19.6\%, 46.3\%, 7.2\% and 24.9\% on the LIP, FIP, GarMoCap and MIG datasets.
It also improves on \projectname{} Body data from the same clips, by a relative 2.1\% to 11.2\%, so the simulated garment adds to what the motion alone provides.
Without real data, the gain over \projectname{} Body is largest on the loose sensors of the MIG dataset, where \projectname{} Garment reaches a macro-F1 of 43.5\% against 37.7\% for \projectname{} Body, a relative improvement of 15.4\% and more than real data from one subject reach (39.8\%).
With all real subjects, the best \projectname{} source remains a relative 1.0\% to 5.5\% above real data alone.

\projectname{} also outperforms existing techniques for generating virtual IMU data.
IMUGPT generates motion from text, as \projectname{} does, but senses it with rigid sensors on the skeleton, so its data describe a body-mounted sensor.
TailorNet adds a garment, but one whose shape follows the pose without any dynamics of its own.
With one real subject, \projectname{} Body + Garment improves on IMUGPT by a relative 19.2\% to 44.9\% and on TailorNet by a relative 3.4\% to 13.7\%.
The gain over TailorNet shows that the benefit comes from simulating how the garment moves, beyond a garment shape that follows the pose.
Without real data, the gap to IMUGPT is largest, with a macro-F1 of 45.6\% on the MIG dataset against 7.2\% for IMUGPT.
Among the \projectname{} sources, \projectname{} Garment and \projectname{} Body + Garment perform similarly. \projectname{} Body + Garment leads without real data on the LIP, FIP and MIG datasets, and \projectname{} Garment leads on the LIP and MIG datasets with two real subjects.
Tables~\ref{tab:har_permodel} and~\ref{tab:har_permodel_b} in Appendix~\ref{app:har} give the results of each model.

\input{tables/pose_main}

\subsection{Pose Estimation}
\label{sec:eval:pose}

\subsubsection{Experimental Setting}
We estimate full-body pose from the garment IMUs of the LIP, FIP and GarMoCap datasets.
We use two networks, an LSTM poser and TransPose~\cite{yi2021transpose}.
The networks are pre-trained on virtual data and then fine-tuned on the real training data, as in sparse inertial pose estimation~\cite{huang2018dip,yi2021transpose,yi2022pip}.
Without real data, the pre-trained network is tested directly.
As in recognition, we standardize the accelerations of each subject with its own statistics, those of the test subject from the earlier half of each recording, and score the later half.
We report the mean angular error over 15 major joints in degrees and the mean joint position error in centimeters.

\subsubsection{Results}

Without real data, \projectname{} data give the lowest joint angle error and the lowest joint position error on all three datasets (Table~\ref{tab:pose_main}).
The best \projectname{} source reduces the angle error relative to IMUGPT by 7.8\% to 11.5\% and relative to TailorNet by 3.2\% to 10.0\%.
Simulating the garment helps most on the FIP and GarMoCap datasets. Relative to \projectname{} Body, \projectname{} Garment reduces the angle error by 8.4\% on FIP and by 10.2\% on GarMoCap. On LIP, \projectname{} Garment and \projectname{} Body give the same error, and \projectname{} Body + Garment is 0.9\% lower.
Pre-training on \projectname{} data also helps with one real subject, reducing the error relative to real data alone by 0.9\% on LIP, 2.2\% on FIP and 5.1\% on GarMoCap.
With two or all real subjects, the real data dominate, and every source is within about 4\% of real data alone.

Tables~\ref{tab:pose_permodel} and~\ref{tab:pose_permodel_pos} in Appendix~\ref{app:pose} give the angle and position errors of each network.

%% file: tables/har_main.tex
\begin{table}[!tbp]
  \caption{Activity recognition performance (macro-F1, \%), averaged over the test subjects and 3 recognition models (per-model results in Appendix~\ref{app:har}). The best result in each block and column is highlighted.}
  \label{tab:har_main}
  \centering
  \small
  \begin{tabular}{@{}lcccc@{}}
    \toprule
    Training data & LIP & FIP & GarMoCap & MIG \\
    \midrule
    \multicolumn{5}{c}{Virtual data only} \\
    \midrule
    IMUGPT & 51.8 $\pm$ 2.3 & 16.2 $\pm$ 1.0 & 24.3 $\pm$ 1.8 & 7.2 $\pm$ 0.2 \\
    TailorNet & 65.2 $\pm$ 2.1 & 20.8 $\pm$ 0.8 & 28.7 $\pm$ 1.2 & 27.5 $\pm$ 1.2 \\
    \projectname{} Body & 65.7 $\pm$ 3.3 & 18.7 $\pm$ 1.7 & \best{29.1\pm0.4} & 37.7 $\pm$ 0.3 \\
    \projectname{} Garment & 58.8 $\pm$ 3.1 & 21.4 $\pm$ 0.9 & 26.6 $\pm$ 1.6 & 43.5 $\pm$ 1.0 \\
    \projectname{} Body + Garment & \best{71.9\pm1.4} & \best{23.1\pm0.6} & 27.6 $\pm$ 1.3 & \best{45.6\pm1.2} \\
    \midrule
    \multicolumn{5}{c}{1 real subject} \\
    \midrule
    Real data only & 66.9 $\pm$ 4.5 & 17.7 $\pm$ 4.3 & 45.8 $\pm$ 2.8 & 39.8 $\pm$ 2.7 \\
    IMUGPT & 66.4 $\pm$ 3.4 & 18.0 $\pm$ 3.4 & 41.2 $\pm$ 1.3 & 34.3 $\pm$ 3.9 \\
    TailorNet & 76.5 $\pm$ 1.5 & 23.9 $\pm$ 1.6 & 47.5 $\pm$ 1.8 & 43.7 $\pm$ 2.7 \\
    \projectname{} Body & 78.2 $\pm$ 4.0 & 23.3 $\pm$ 2.4 & 47.2 $\pm$ 0.9 & 48.7 $\pm$ 2.1 \\
    \projectname{} Garment & 77.8 $\pm$ 0.9 & 25.6 $\pm$ 2.1 & 47.9 $\pm$ 1.6 & \best{49.8\pm3.3} \\
    \projectname{} Body + Garment & \best{80.0\pm1.3} & \best{25.9\pm0.9} & \best{49.1\pm1.7} & 49.7 $\pm$ 6.1 \\
    \midrule
    \multicolumn{5}{c}{2 real subjects} \\
    \midrule
    Real data only & 80.1 $\pm$ 3.0 & 29.0 $\pm$ 4.0 & 56.6 $\pm$ 2.6 & 49.8 $\pm$ 2.2 \\
    IMUGPT & 78.5 $\pm$ 2.8 & 26.9 $\pm$ 3.5 & 54.5 $\pm$ 1.0 & 46.9 $\pm$ 2.5 \\
    TailorNet & 82.2 $\pm$ 1.0 & 32.6 $\pm$ 0.6 & 56.8 $\pm$ 1.2 & 50.9 $\pm$ 2.1 \\
    \projectname{} Body & 84.7 $\pm$ 0.7 & 33.1 $\pm$ 1.8 & 56.8 $\pm$ 1.2 & 52.6 $\pm$ 1.9 \\
    \projectname{} Garment & \best{85.0\pm1.5} & 35.2 $\pm$ 2.4 & 58.1 $\pm$ 1.3 & \best{57.3\pm0.6} \\
    \projectname{} Body + Garment & 84.8 $\pm$ 1.5 & \best{35.5\pm0.8} & \best{58.6\pm1.9} & 56.4 $\pm$ 0.9 \\
    \midrule
    \multicolumn{5}{c}{All real subjects} \\
    \midrule
    Real data only & 87.5 $\pm$ 0.3 & 55.7 $\pm$ 0.6 & 74.8 $\pm$ 0.9 & 67.3 $\pm$ 0.6 \\
    IMUGPT & 87.3 $\pm$ 1.0 & 54.1 $\pm$ 1.2 & 74.5 $\pm$ 0.8 & 65.4 $\pm$ 1.7 \\
    TailorNet & 88.2 $\pm$ 0.8 & 55.9 $\pm$ 0.4 & 74.9 $\pm$ 0.6 & 68.1 $\pm$ 0.8 \\
    \projectname{} Body & 88.3 $\pm$ 0.9 & 54.0 $\pm$ 0.8 & 74.0 $\pm$ 0.8 & 67.7 $\pm$ 1.1 \\
    \projectname{} Garment & \best{88.4\pm1.1} & 56.6 $\pm$ 0.6 & \best{76.8\pm1.8} & \best{71.0\pm1.4} \\
    \projectname{} Body + Garment & 88.2 $\pm$ 1.5 & \best{57.2\pm0.5} & 75.7 $\pm$ 1.4 & 70.6 $\pm$ 1.8 \\
    \bottomrule
  \end{tabular}
\end{table}

%% file: tables/pose_main.tex
\begin{table}[!tbp]
  \caption{Pose estimation performance: mean angular error over 15 major joints (degrees) and mean joint position error (cm), lower is better, averaged over 2 pose networks. The best result in each block and column is highlighted.}
  \label{tab:pose_main}
  \centering
  \footnotesize
  \setlength{\tabcolsep}{3pt}
  \begin{tabular}{@{}lcccccc@{}}
    \toprule
    & \multicolumn{3}{c}{Angular error ($^\circ$)} & \multicolumn{3}{c}{Position error (cm)} \\
    \cmidrule(lr){2-4}\cmidrule(l){5-7}
    Training data & LIP & FIP & GarMoCap & LIP & FIP & GarMoCap \\
    \midrule
    \multicolumn{7}{c}{Virtual data only} \\
    \midrule
    IMUGPT & 31.4 $\pm$ 0.48 & 26.7 $\pm$ 0.31 & 22.9 $\pm$ 0.17 & 11.3 $\pm$ 0.23 & 8.6 $\pm$ 0.08 & 10.8 $\pm$ 0.08 \\
    TailorNet & 29.3 $\pm$ 0.27 & 25.5 $\pm$ 0.62 & 23.4 $\pm$ 0.31 & 10.5 $\pm$ 0.14 & 8.6 $\pm$ 0.15 & 10.5 $\pm$ 0.19 \\
    \projectname{} Body & 28.6 $\pm$ 0.28 & 25.8 $\pm$ 0.21 & 23.5 $\pm$ 0.27 & 10.3 $\pm$ 0.08 & 8.8 $\pm$ 0.11 & 10.3 $\pm$ 0.18 \\
    \projectname{} Garment & 28.6 $\pm$ 0.47 & \best{23.7\pm0.39} & \best{21.1\pm0.18} & \best{9.7\pm0.11} & \best{8.0\pm0.24} & 9.5 $\pm$ 0.15 \\
    \projectname{} Body + Garment & \best{28.3\pm0.31} & 24.4 $\pm$ 0.45 & 21.1 $\pm$ 0.08 & 9.9 $\pm$ 0.06 & 8.2 $\pm$ 0.13 & \best{9.5\pm0.10} \\
    \midrule
    \multicolumn{7}{c}{1 real subject} \\
    \midrule
    Real data only & 22.9 $\pm$ 1.28 & 15.1 $\pm$ 1.04 & 13.9 $\pm$ 0.59 & 7.7 $\pm$ 0.37 & 4.8 $\pm$ 0.43 & 6.8 $\pm$ 0.34 \\
    IMUGPT & 22.7 $\pm$ 1.22 & 15.3 $\pm$ 1.00 & 13.7 $\pm$ 0.46 & 7.7 $\pm$ 0.37 & 4.9 $\pm$ 0.44 & 6.7 $\pm$ 0.25 \\
    TailorNet & \best{22.7\pm1.17} & 15.1 $\pm$ 0.94 & 13.7 $\pm$ 0.50 & \best{7.7\pm0.38} & 4.9 $\pm$ 0.40 & 6.7 $\pm$ 0.35 \\
    \projectname{} Body & 22.7 $\pm$ 1.13 & 15.4 $\pm$ 1.29 & 13.7 $\pm$ 0.47 & 7.7 $\pm$ 0.38 & 5.0 $\pm$ 0.57 & 6.7 $\pm$ 0.33 \\
    \projectname{} Garment & 22.9 $\pm$ 1.26 & \best{14.7\pm0.68} & 13.4 $\pm$ 0.54 & 7.8 $\pm$ 0.37 & \best{4.7\pm0.30} & 6.5 $\pm$ 0.31 \\
    \projectname{} Body + Garment & 22.8 $\pm$ 1.33 & 14.9 $\pm$ 0.80 & \best{13.2\pm0.48} & 7.7 $\pm$ 0.45 & 4.8 $\pm$ 0.37 & \best{6.4\pm0.30} \\
    \midrule
    \multicolumn{7}{c}{2 real subjects} \\
    \midrule
    Real data only & 20.7 $\pm$ 0.57 & 13.9 $\pm$ 0.27 & 12.4 $\pm$ 0.19 & 7.0 $\pm$ 0.10 & 4.3 $\pm$ 0.12 & 6.0 $\pm$ 0.06 \\
    IMUGPT & 20.5 $\pm$ 0.37 & 13.9 $\pm$ 0.30 & 12.3 $\pm$ 0.18 & 7.0 $\pm$ 0.03 & 4.3 $\pm$ 0.15 & 5.9 $\pm$ 0.05 \\
    TailorNet & 20.5 $\pm$ 0.44 & 13.9 $\pm$ 0.38 & 12.2 $\pm$ 0.07 & \best{7.0\pm0.11} & 4.3 $\pm$ 0.13 & 5.9 $\pm$ 0.11 \\
    \projectname{} Body & 20.6 $\pm$ 0.38 & 13.8 $\pm$ 0.30 & 12.2 $\pm$ 0.17 & 7.1 $\pm$ 0.04 & 4.3 $\pm$ 0.14 & 5.9 $\pm$ 0.07 \\
    \projectname{} Garment & \best{20.5\pm0.37} & \best{13.7\pm0.28} & 12.3 $\pm$ 0.27 & 7.0 $\pm$ 0.09 & 4.3 $\pm$ 0.10 & 5.9 $\pm$ 0.11 \\
    \projectname{} Body + Garment & 20.6 $\pm$ 0.38 & 13.7 $\pm$ 0.35 & \best{12.0\pm0.14} & 7.0 $\pm$ 0.05 & \best{4.2\pm0.15} & \best{5.8\pm0.11} \\
    \midrule
    \multicolumn{7}{c}{All real subjects} \\
    \midrule
    Real data only & 19.1 $\pm$ 0.09 & 12.5 $\pm$ 0.05 & 10.5 $\pm$ 0.08 & \best{6.5\pm0.04} & 3.8 $\pm$ 0.01 & 5.1 $\pm$ 0.02 \\
    IMUGPT & 19.1 $\pm$ 0.11 & 12.4 $\pm$ 0.05 & 10.7 $\pm$ 0.09 & 6.6 $\pm$ 0.03 & 3.8 $\pm$ 0.02 & 5.2 $\pm$ 0.03 \\
    TailorNet & 19.2 $\pm$ 0.01 & 12.3 $\pm$ 0.06 & 10.7 $\pm$ 0.02 & 6.5 $\pm$ 0.02 & 3.8 $\pm$ 0.02 & 5.2 $\pm$ 0.02 \\
    \projectname{} Body & \best{19.1\pm0.01} & 12.3 $\pm$ 0.07 & 10.6 $\pm$ 0.01 & 6.5 $\pm$ 0.01 & 3.8 $\pm$ 0.02 & 5.1 $\pm$ 0.01 \\
    \projectname{} Garment & 19.1 $\pm$ 0.04 & \best{12.2\pm0.05} & 10.6 $\pm$ 0.07 & 6.5 $\pm$ 0.04 & 3.8 $\pm$ 0.02 & 5.1 $\pm$ 0.03 \\
    \projectname{} Body + Garment & 19.1 $\pm$ 0.01 & 12.2 $\pm$ 0.05 & \best{10.4\pm0.03} & 6.6 $\pm$ 0.03 & \best{3.7\pm0.03} & \best{5.0\pm0.01} \\
    \bottomrule
  \end{tabular}
\end{table}

%% file: sections/discussion/main.tex
\section{Discussion}
\label{sec:discussion}

The evaluation shows that simulated garment data improve activity recognition and pose estimation on real garments, beyond real data alone and beyond virtual data from body-mounted sensors (IMUGPT, \projectname{} Body) or from a learned garment model without dynamics (TailorNet).
The benefit is largest when real data are scarce.
In this section we first analyze when and why garment data help (Section~\ref{sec:disc:why}).
We then ask how the benefit changes when the garment is simulated on recorded instead of generated motion (Section~\ref{sec:eval:realmotion}) and when it is simulated on a garment other than the deployed one (Section~\ref{sec:disc:matching}).
Finally, we summarize what these findings mean for practitioners (Section~\ref{sec:disc:practitioners}) and discuss limitations (Section~\ref{sec:disc:limitations}).

\subsection{When and Why Do Garment Data Help?}
\label{sec:disc:why}

\subsubsection{How Much Virtual Data Is Needed?}
\label{sec:disc:amount}
\projectname{} can generate virtual data in any amount, at the cost of simulation time. A practical question is therefore how much to generate, and whether generating more would still pay off.
We study this on activity recognition on the MIG dataset, with the setting of Section~\ref{sec:eval:har}.
We train on 25\%, 50\% or all of the virtual clips, that is, about 22, 45 or 89 minutes of virtual data, with 0, 1, 2 or all real training subjects.
The smaller sets are drawn per activity, so each keeps the activity mix of the full set.
The setting without real data is the one \projectname{} ultimately aims at, a new garment for which nothing has been recorded. The settings with real data show whether generating more virtual data still pays off once some recordings exist.

We find that more virtual data help most without real data (Table~\ref{tab:har_amount}). There, going from 25\% to 100\% of the clips improves macro-F1 by a relative 22.0\% for \projectname{} Body, 19.8\% for \projectname{} Garment and 17.8\% for \projectname{} Body + Garment data, and every source improves at each step.
With one or two real subjects, more virtual data still help the garment sources, by up to 4.4\% relative, and with all real subjects the three amounts perform alike.
Without real data, the gains have not leveled off at the largest set, so generating more virtual data could improve recognition further.
We therefore expect generation at a larger scale, with more descriptions per activity and more varied motions, to raise performance further, and we see it as a promising direction for future work.

\input{tables/har_amount}

\subsubsection{Tight or Loose Sensors?}
\label{sec:disc:tightloose}

A sensor on a loose part of a garment moves more relative to the body than one where the fabric is pressed against the body.
One might therefore expect garment simulation to matter only for loose sensors, and body data to suffice for tight ones.
We test this on the MIG dataset, whose jacket and trousers carry sensors at both kinds of positions.
The main MIG results use six loose-fitting sensors, where the fabric hangs free (Section~\ref{sec:eval:har}).
Here we compare them with a tight set of six sensors in comparable regions, where the garment is pressed against the body: the forearm, upper arm, chest, waist, thigh and shank.
Both sets are evaluated with the setting of Section~\ref{sec:eval:har}, with every virtual data source and 0, 1, 2 or all real training subjects.
Table~\ref{tab:tightloose} shows the results.

On both sets, \projectname{} Garment data outperform \projectname{} Body data at every amount of real data.
The gain is larger on the tight set when real data are scarce. Without real data, \projectname{} Garment improves on \projectname{} Body by a relative 59.5\% on the tight set against 15.4\% on the loose set, and with one real subject by 10.0\% against 2.3\%.
With two or all real subjects, the loose set gains slightly more, by 8.9\% and 4.9\% on the loose set against 7.5\% and 4.3\% on the tight set.
On the tight set, \projectname{} Body + Garment data are the best source at every amount of real data, and \projectname{} Garment data are within 0.8 points of it.
Simulating the garment therefore pays off for sensors on tight-fitting as well as loose-fitting parts of a garment, and Section~\ref{sec:disc:placement} examines where on the garment it matters most.

\input{tables/tightloose}

\subsubsection{Where on a Garment Does Physics Matter?}
\label{sec:disc:placement}
A densely instrumented simulated garment lets us ask where on a garment simulating the garment makes a difference.
Sensor placement has been studied on the real MIG data with deep models~\cite{enokibori2025positions} and in simulation for body-mounted sensors on motion-capture data~\cite{ray2025w2w}, but not for the difference between garment and body sensors.
We trained single-sensor recognition models at each of the 396 MIG sensor positions, once on simulated garment data and once on body data from the same position, without real training data, and tested them on the real garment data of the held-out subjects.
We used the three recognition models of Section~\ref{sec:eval:har}.
Each model is evaluated with the protocol of Section~\ref{sec:eval:har} over all 12 leave-one-subject-out folds.

Figure~\ref{fig:site_map} shows that garment physics helps at most positions.
Garment data beat body data at 255, 264 and 212 of the 396 positions in the three models.
Averaged over all positions, they improve macro-F1 by a relative 11.1\%, 14.4\% and 4.8\%.
The gain holds in every model on the sleeves and the thighs.
It is largest on the forearm sleeves, by a relative 59.5\% to 105.7\%, where garment data win at 47 or 48 of the 48 positions.
There the sleeve turns around the arm and slides along it, and body data cannot reproduce this motion.
On the jacket trunk, garment and body data are close ($-8.3\%$ to $+3.2\%$).
Body data remain better on the lower trouser legs in every model ($-4.6\%$ to $-18.8\%$). 
We also combined the sensors of the tight and the loose set of Section~\ref{sec:disc:tightloose}, using one to six sensors of each set at once with the same protocol and no real training data. For two to five sensors, we averaged over several subsets of the set (Table~\ref{tab:sensorcount}).
The advantage of garment data over body data grows as sensors are combined.
On the tight set, it rises from a relative 20.7\% with one sensor to 56.2\% with all six.
Here garment data improve from 31.6\% to 51.0\% macro-F1 as sensors are added, whereas body data gain little (26.2\% to 32.7\%).
On the loose set, the advantage grows from 5.2\% with one sensor to between 14.0\% and 18.7\% with two to six sensors.
Simulating the garment therefore matters even more for garments that carry several sensors than for a single sensor.

\input{tables/sensorcount}

\input{figure/site_map/site_map}

\subsection{Recorded Instead of Generated Motion}
\label{sec:eval:realmotion}
Generated motion differs from real recordings, so part of the remaining error may come from the motion rather than from the garment.
To separate the two, we simulate the garment on recorded motion instead of generated motion.
We use GarMoCap~\cite{garmocap_gid}, the one dataset that records the wearer's full-body motion together with loose-garment IMUs on both a jacket and trousers.
We simulate the matched GarMoCap garment on the recorded motion of every subject.
In every leave-one-subject-out fold, the virtual data come only from the recorded motion of that fold's training subjects, so the test subject's motion is never simulated for training.
For a fair comparison, we draw generated motion at random from the generated data of Section~\ref{sec:eval}, simulated on the same garment, so that the virtual data from recorded and from generated motion are of the same size.
Training and evaluation follow the setting of Sections~\ref{sec:eval:har} and~\ref{sec:eval:pose}.
Table~\ref{tab:realmotion} shows the results.

Recorded motion improves every virtual source when real data are scarce, most of all without real data.
Without real data, it lowers the joint angle error relative to generated motion by 29.7\% for \projectname{} Body, 30.5\% for \projectname{} Garment and 33.3\% for \projectname{} Body + Garment data, and it raises recognition macro-F1 by 16.4\%, 23.9\% and 20.1\%. Both pose networks show this drop, by 28.4\% to 32.9\% for the LSTM poser and by 31.0\% to 33.7\% for TransPose.
With one real subject, \projectname{} Body + Garment data from recorded motion raise macro-F1 from 45.8\% with real data alone to 56.5\%, against 49.2\% with generated motion, and lower the joint angle error from 13.9$^\circ$ to 12.9$^\circ$, against 13.4$^\circ$.
\projectname{} Body + Garment data from recorded motion give the lowest pose error at every amount of real data.
With all real subjects, recorded and generated motion perform alike.
Garment data outperform body data with either motion. Without real data, \projectname{} Garment lowers the joint angle error relative to \projectname{} Body by 12.1\% with recorded and by 11.1\% with generated motion.
The gains from realistic motion and from garment simulation therefore add up, and more realistic and more varied generated motion is a direct way to improve \projectname{} further (Section~\ref{sec:disc:limitations}).

\input{tables/realmotion}

\subsection{Matched or Generic Garment}
\label{sec:disc:matching}
So far, each dataset was simulated with its matched garment, recreated from photographs of the garment the participants wore.
Generating the garment of a deployment from a photograph or a description is a capability that learned garment models such as TailorNet lack, as they are limited to the few garment templates they were trained on.
Here we ask how much this matching is worth, by comparing the matched garment with a different garment of the same type.
Every garment in \projectname{} carries the sensor positions of all datasets, so the same sensors can be read on a different garment.
We simulate the recorded GarMoCap clips of Section~\ref{sec:eval:realmotion} on the matched GarMoCap garment and on the MIG jacket and trousers.
We then train and evaluate the two pose networks and the three recognition models on \projectname{} Garment data from each garment, as in Section~\ref{sec:eval:realmotion}.
Table~\ref{tab:matching} shows the results.

For pose estimation, the matched garment gives a lower joint angle error than the MIG garment at every amount of real data. The difference is largest without real data, 1.7\%, and it appears with both pose networks. Without real data, both garments lower the error relative to body data from the same recorded motion by 10\% to 12\% (Table~\ref{tab:realmotion}).
For recognition, the matched garment helps most without real data, where it improves macro-F1 by 14.9\%, from 28.1\% to 32.3\%. With real data, it stays ahead of the MIG garment by up to 1.0 point.
The matched garment gives the best result in every setting, tied with real data alone for pose estimation with all real subjects.
Matching the garment to the deployment therefore pays off most when no real data exist, which is exactly when a new smart garment is introduced.
At the same time, a different garment of the right type still keeps the benefit of garment simulation over body data, so \projectname{} remains useful when the exact garment is not known.

\input{tables/matching}

\subsection{Guidelines for Practitioners}
\label{sec:disc:practitioners}
Consider two teams who want to use inertial sensors built into clothing.
The first is developing a new smart garment, for example a workout shirt or a jacket with sensors in the sleeves, and needs an activity recognition model before the product reaches its users.
The second wants to monitor patients in everyday clothes, for example the daily activity of older adults or the exercises of rehabilitation patients, without asking them to strap on sensors.
Both face the same problem.
A model for a garment-mounted sensor needs training data recorded with that garment, at those sensor positions, and recording such data means recruiting participants, dressing them in the garment and labeling what they do.
For a new product this has to be repeated for every cut and sensor layout, and for patients every recording session is a burden.
Existing virtual IMU data do not solve this problem, because they describe sensors strapped to the body.

\projectname{} offers these teams a different starting point.
From a photograph or a description of their garment, it generates a matching garment, places the virtual sensors where the real ones will sit and produces training data for the activities they describe in text.
Our results show how to make the most of these data.
\begin{itemize}
  \item \textbf{Start before any recording.} Without real data, \projectname{} data already outperform the body-mounted virtual data of IMUGPT and the learned garment model TailorNet on every dataset, for recognition and for pose estimation (Sections~\ref{sec:eval:har} and~\ref{sec:eval:pose}). A team can therefore train a first model before recruiting a single participant.
  \item \textbf{Record a few people.} The gains are largest with one or two real subjects, the situation of a pilot study or a small patient cohort. For recognition, \projectname{} Garment and \projectname{} Body + Garment data work best, and pre-training on \projectname{} data before fine-tuning on the real recordings helps pose estimation (Sections~\ref{sec:eval:har} and~\ref{sec:eval:pose}). For recognition, virtual data still add a small gain even with all real subjects.
  \item \textbf{Simulate the garment that will be worn.} Generating the deployed garment instead of a different one of the same type helps most when no real data exist (Section~\ref{sec:disc:matching}), so the photograph of the product or of the patients' usual clothing is worth providing.
  \item \textbf{Use several sensors.} The advantage of garment data over body data grows when sensors are combined (Section~\ref{sec:disc:placement}), so simulation is most valuable for garments that carry several sensors.
  \item \textbf{Sensors on tight and on loose fabric both benefit} (Section~\ref{sec:disc:tightloose}), so a designer does not need to restrict sensors to tight-fitting parts to profit from simulation. The forearm sleeves gain most from physics (Section~\ref{sec:disc:placement}).
  \item \textbf{Generate more, and use real motion when it exists.} More virtual data keep improving recognition, most of all before real data are available (Section~\ref{sec:disc:amount}). If the team can record the wearers' motion, for example with a camera-based motion capture of the target exercises, simulating the garment on this motion adds to the benefit of the garment simulation (Section~\ref{sec:eval:realmotion}).
\end{itemize}
In both scenarios, \projectname{} thus turns a garment design and a list of activities into a working model with little or no real data, and makes each real recording that is collected count for more.

\subsection{Limitations and Future Work}
\label{sec:disc:limitations}

\paragraph{Scale and diversity of the virtual data.}
Our virtual datasets are still small. Each dataset receives 20 generated descriptions per activity, all simulated on a single body shape, the average SMPL body ($\beta=0$).
Three of our findings suggest that larger and more varied virtual data would increase the benefit of \projectname{}.
First, without real data, more virtual data still improve recognition at the largest amount we generated (Section~\ref{sec:disc:amount}).
Second, recorded motion lowers the pose error of every virtual source by more than a quarter without real data (Section~\ref{sec:eval:realmotion}). Generated motion still differs from real recordings and does not yet span the variety of ways in which the recorded participants move.
Third, virtual data help most when real data come from only one or two people (Section~\ref{sec:eval:har}), where they add variation between people that the real data lack. Real wearers also differ in build, and the same garment fits them differently. A jacket that is loose on a slim person can be tight on a broader one, which changes how far the fabric moves relative to the body.
Generating more descriptions per activity, adopting better motion generators as they appear, and simulating a range of body shapes, each dressed in the garment at its own size or at a fixed size, are therefore natural next steps. 

\paragraph{Designing smart garments in simulation.}
Because a garment in \projectname{} is a sewing pattern, its cut and its sensor layout can be changed and simulated before the garment is produced, so candidate designs for a planned smart garment can be compared before any participant is recorded.
Garment design in simulation is already established in computer graphics, where sewing patterns are generated programmatically~\cite{garmentcode} and optimized with differentiable cloth simulation~\cite{li2024diffavatar}, and in wearable computing, simulation has begun to guide where IMUs are placed on the body~\cite{ray2025w2w} and where textile energy harvesters are placed on garments~\cite{ray2026spotteg}.
Optimizing the cut of a garment together with its sensor layout for a downstream task, such as recognizing a given set of activities, has not been explored, and \projectname{} provides the data needed to do so.

\paragraph{Further sensors and modalities.}
A smartphone carried in a trouser pocket is itself a garment-mounted sensor. Many smartphone HAR datasets were recorded this way, and simulating a phone inside a pocket, as a rigid body moving within the fabric, would extend \projectname{} to them.
The simulation also describes the whole garment, not only the points where IMUs sit. It gives the stretch and bending of the fabric everywhere, which is what textile strain and pressure sensors measure, so the same simulations could provide virtual data for these sensors as well.
Finally, the generated clips carry the text descriptions they were generated from, so they form IMU--text pairs. Sensor--language models are currently trained on body-mounted virtual data~\cite{mosaic2026}, and pre-training them on garment-mounted data could carry their zero-shot recognition over to smart clothing.

%% file: tables/har_amount.tex
\begin{table}[!htbp]
  \caption{Activity recognition on MIG (macro-F1, \%) with 25, 50 or 100\% of the virtual clips, for each amount of real training data. The best result in each column is highlighted.}
  \label{tab:har_amount}
  \centering
  \footnotesize
  \setlength{\tabcolsep}{3pt}
  \begin{tabular}{@{}llcccc@{}}
    \toprule
    & & \multicolumn{4}{c}{Real training subjects} \\
    \cmidrule(l){3-6}
    Virtual data & Amount & None & 1 subject & 2 subjects & All subjects \\
    \midrule
    Real data only & -- & -- & 39.8 $\pm$ 2.7 & 49.8 $\pm$ 2.2 & 67.3 $\pm$ 0.6 \\
    \midrule
    \projectname{} Body & 25\% & 30.9 $\pm$ 2.2 & 47.1 $\pm$ 1.0 & 53.6 $\pm$ 0.8 & 68.8 $\pm$ 0.7 \\
     & 50\% & 33.0 $\pm$ 0.5 & 45.9 $\pm$ 1.7 & 52.4 $\pm$ 2.1 & 69.0 $\pm$ 0.2 \\
     & 100\% & 37.7 $\pm$ 0.3 & 48.7 $\pm$ 2.1 & 52.6 $\pm$ 1.9 & 67.7 $\pm$ 1.1 \\
    \midrule
    \projectname{} Garment & 25\% & 36.3 $\pm$ 2.4 & 47.7 $\pm$ 3.7 & 55.7 $\pm$ 0.8 & \best{71.0\pm0.6} \\
     & 50\% & 42.2 $\pm$ 0.9 & 47.5 $\pm$ 2.6 & 54.7 $\pm$ 1.1 & 70.9 $\pm$ 1.5 \\
     & 100\% & 43.5 $\pm$ 1.0 & \best{49.8\pm3.3} & \best{57.3\pm0.6} & 71.0 $\pm$ 1.4 \\
    \midrule
    \projectname{} Body + Garment & 25\% & 38.7 $\pm$ 1.1 & 47.7 $\pm$ 3.3 & 56.1 $\pm$ 3.5 & 70.1 $\pm$ 0.4 \\
     & 50\% & 42.9 $\pm$ 2.0 & 48.6 $\pm$ 4.0 & 54.9 $\pm$ 1.6 & 69.5 $\pm$ 0.5 \\
     & 100\% & \best{45.6\pm1.2} & 49.7 $\pm$ 6.1 & 56.4 $\pm$ 0.9 & 70.6 $\pm$ 1.8 \\
    \bottomrule
  \end{tabular}
\end{table}

%% file: tables/tightloose.tex
\begin{table}[!tbp]
  \caption{Activity recognition on MIG with the tight and the loose sensor set (macro-F1, \%), for each amount of real training data. The best result in each column is highlighted.}
  \label{tab:tightloose}
  \centering
  \scriptsize
  \setlength{\tabcolsep}{2pt}
  \begin{tabular}{@{}lcccccccc@{}}
    \toprule
    & \multicolumn{4}{c}{Tight sensor set} & \multicolumn{4}{c}{Loose sensor set} \\
    \cmidrule(lr){2-5} \cmidrule(l){6-9}
    Training data & None & 1 subject & 2 subjects & All subjects & None & 1 subject & 2 subjects & All subjects \\
    \midrule
    Real data only & -- & 42.4 $\pm$ 3.0 & 53.5 $\pm$ 2.9 & 66.8 $\pm$ 0.2 & -- & 39.8 $\pm$ 2.7 & 49.8 $\pm$ 2.2 & 67.3 $\pm$ 0.6 \\
    IMUGPT & 6.7 $\pm$ 0.5 & 43.5 $\pm$ 1.2 & 54.6 $\pm$ 2.7 & 67.1 $\pm$ 0.6 & 7.2 $\pm$ 0.2 & 34.3 $\pm$ 3.9 & 46.9 $\pm$ 2.5 & 65.4 $\pm$ 1.7 \\
    TailorNet & 26.7 $\pm$ 2.9 & 48.2 $\pm$ 1.4 & 54.9 $\pm$ 0.4 & 66.7 $\pm$ 1.0 & 27.5 $\pm$ 1.2 & 43.7 $\pm$ 2.7 & 50.9 $\pm$ 2.1 & 68.1 $\pm$ 0.8 \\
    \projectname{} Body & 32.6 $\pm$ 1.9 & 49.0 $\pm$ 0.4 & 55.8 $\pm$ 2.6 & 67.0 $\pm$ 2.0 & 37.7 $\pm$ 0.3 & 48.7 $\pm$ 2.1 & 52.6 $\pm$ 1.9 & 67.7 $\pm$ 1.1 \\
    \projectname{} Garment & 52.0 $\pm$ 1.0 & 53.9 $\pm$ 2.7 & 60.0 $\pm$ 0.6 & 69.9 $\pm$ 0.6 & 43.5 $\pm$ 1.0 & \best{49.8\pm3.3} & \best{57.3\pm0.6} & \best{71.0\pm1.4} \\
    \projectname{} Body + Garment & \best{52.8\pm0.7} & \best{54.6\pm1.7} & \best{60.8\pm2.9} & \best{70.6\pm0.9} & \best{45.6\pm1.2} & 49.7 $\pm$ 6.1 & 56.4 $\pm$ 0.9 & 70.6 $\pm$ 1.8 \\
    \bottomrule
  \end{tabular}
\end{table}

%% file: tables/sensorcount.tex
\begin{table}[!tbp]
  \caption{Activity recognition on MIG without real training data, with 1 to 6 sensors drawn from the tight and the loose sensor set (macro-F1, \%, mean over the three recognition models and over the sensor subsets of each size). Improvement over Body is the relative improvement of \projectname{} Garment over \projectname{} Body.}
  \label{tab:sensorcount}
  \centering
  \small
  \begin{tabular}{@{}ccccccc@{}}
    \toprule
    & \multicolumn{3}{c}{Tight set} & \multicolumn{3}{c}{Loose set} \\
    \cmidrule(lr){2-4}\cmidrule(lr){5-7}
    Sensors & Body & Garment & Improvement over Body & Body & Garment & Improvement over Body \\
    \midrule
    1 & 26.2 $\pm$ 0.7 & 31.6 $\pm$ 0.2 & +20.7\% & 25.2 $\pm$ 0.2 & 26.5 $\pm$ 0.2 & +5.2\% \\
    2 & 30.3 $\pm$ 0.4 & 42.1 $\pm$ 0.3 & +38.9\% & 28.5 $\pm$ 0.7 & 32.5 $\pm$ 0.2 & +14.0\% \\
    3 & 30.5 $\pm$ 0.7 & 42.0 $\pm$ 0.4 & +37.5\% & 32.0 $\pm$ 0.8 & 37.7 $\pm$ 0.7 & +17.7\% \\
    4 & 31.6 $\pm$ 0.9 & 45.7 $\pm$ 0.5 & +44.6\% & 33.0 $\pm$ 0.4 & 39.1 $\pm$ 0.5 & +18.7\% \\
    5 & 31.2 $\pm$ 0.8 & 48.2 $\pm$ 0.8 & +54.9\% & 34.5 $\pm$ 1.3 & 40.3 $\pm$ 0.9 & +16.9\% \\
    6 & 32.7 $\pm$ 0.7 & 51.0 $\pm$ 1.6 & +56.2\% & 37.5 $\pm$ 1.1 & 44.1 $\pm$ 1.6 & +18.0\% \\
    \bottomrule
  \end{tabular}
\end{table}

%% file: figure/site_map/site_map.tex
\begin{figure}[!tbp]
  \centering
  \includegraphics[width=\linewidth]{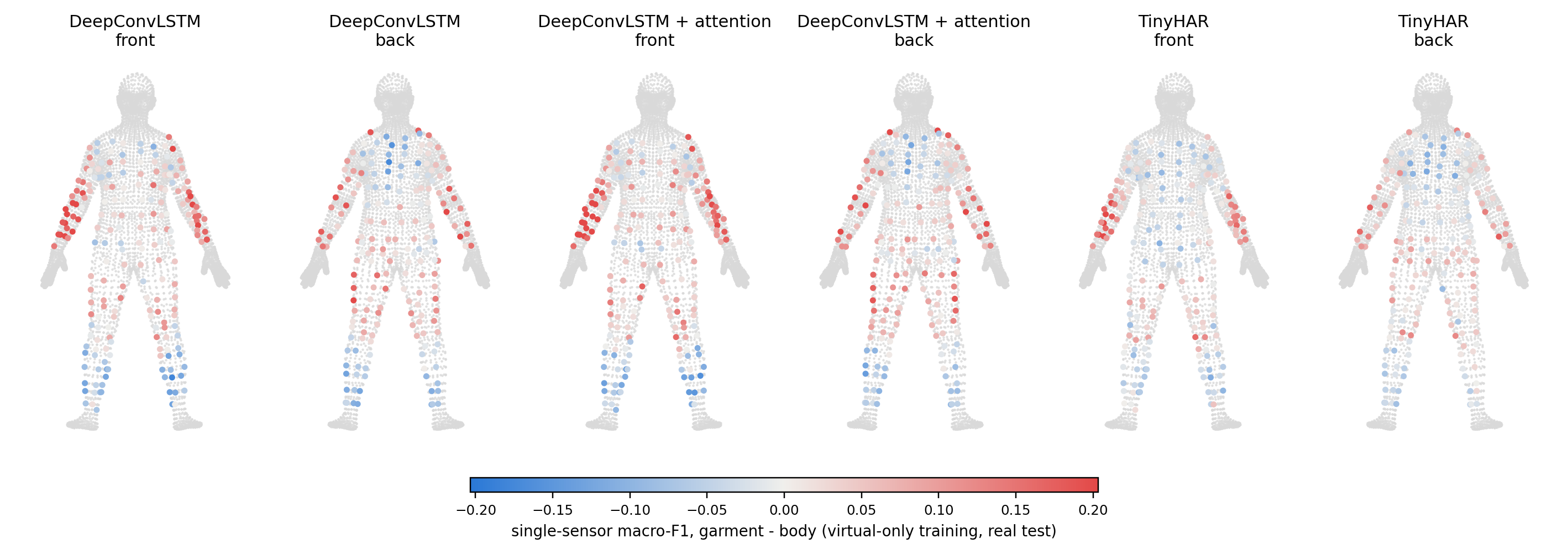}
  \caption{Per-position difference in single-sensor macro-F1 between training on simulated garment IMU data and on body IMU data from the same position (virtual data only, tested on the real MIG data). Red: garment data better; blue: body data better. The front view faces the wearer. }
  \label{fig:site_map}
\end{figure}

%% file: tables/realmotion.tex
\begin{table}[!tbp]
  \caption{GarMoCap pose estimation and activity recognition when the garment is simulated on the dataset's \emph{recorded} training motion or on the same amount of generated motion, both simulated on the matched garment.  The best result in each column is highlighted.}
  \label{tab:realmotion}
  \centering
  \scriptsize
  \setlength{\tabcolsep}{2pt}
  \begin{tabular}{@{}llcccccccc@{}}
    \toprule
    & & \multicolumn{4}{c}{Pose: joint angle error ($^\circ$)} & \multicolumn{4}{c}{Recognition: macro-F1 (\%)} \\
    \cmidrule(lr){3-6} \cmidrule(l){7-10}
    & Training data & None & 1 subject & 2 subjects & All & None & 1 subject & 2 subjects & All \\
    \midrule
     & Real data only & -- & 13.9 $\pm$ 0.59 & 12.4 $\pm$ 0.19 & 10.5 $\pm$ 0.08 & -- & 45.8 $\pm$ 2.8 & 56.6 $\pm$ 2.6 & 74.8 $\pm$ 0.9 \\
    \midrule
    Recorded & \projectname{} Body & 16.8 $\pm$ 0.04 & 13.3 $\pm$ 0.41 & 12.1 $\pm$ 0.12 & 10.5 $\pm$ 0.06 & 29.1 $\pm$ 1.6 & 52.7 $\pm$ 0.5 & 61.4 $\pm$ 0.8 & 76.3 $\pm$ 0.9 \\
     & \projectname{} Garment & 14.8 $\pm$ 0.12 & 13.0 $\pm$ 0.65 & 12.0 $\pm$ 0.30 & 10.5 $\pm$ 0.05 & 32.1 $\pm$ 2.2 & 55.3 $\pm$ 2.7 & 63.7 $\pm$ 1.1 & \best{77.4\pm1.1} \\
     & \projectname{} Body + Garment & \best{14.3\pm0.11} & \best{12.9\pm0.60} & \best{11.9\pm0.29} & \best{10.4\pm0.05} & \best{32.9\pm0.4} & \best{56.5\pm1.6} & \best{63.8\pm1.4} & 76.0 $\pm$ 0.8 \\
    \midrule
    Generated & \projectname{} Body & 23.9 $\pm$ 0.09 & 13.9 $\pm$ 0.31 & 12.4 $\pm$ 0.11 & 10.7 $\pm$ 0.10 & 25.0 $\pm$ 0.6 & 47.0 $\pm$ 0.8 & 57.9 $\pm$ 1.3 & 76.0 $\pm$ 1.1 \\
     & \projectname{} Garment & 21.2 $\pm$ 0.10 & 13.5 $\pm$ 0.57 & 12.3 $\pm$ 0.27 & 10.5 $\pm$ 0.03 & 25.9 $\pm$ 2.4 & 48.9 $\pm$ 0.6 & 58.7 $\pm$ 1.6 & 76.4 $\pm$ 1.4 \\
     & \projectname{} Body + Garment & 21.5 $\pm$ 0.23 & 13.4 $\pm$ 0.59 & 12.1 $\pm$ 0.17 & 10.5 $\pm$ 0.07 & 27.4 $\pm$ 1.1 & 49.2 $\pm$ 0.3 & 59.9 $\pm$ 0.4 & 76.8 $\pm$ 0.1 \\
    \bottomrule
  \end{tabular}
\end{table}

%% file: tables/matching.tex
\begin{table}[!tbp]
  \caption{GarMoCap pose estimation and activity recognition with \projectname{} Garment data from the matched GarMoCap garment and from the generic MIG garment, both simulated on the same recorded motion clips. The best result in each column is highlighted.}
  \label{tab:matching}
  \centering
  \scriptsize
  \setlength{\tabcolsep}{2pt}
  \begin{tabular}{@{}llcccccccc@{}}
    \toprule
    & & \multicolumn{4}{c}{Pose: joint angle error ($^\circ$)} & \multicolumn{4}{c}{Recognition: macro-F1 (\%)} \\
    \cmidrule(lr){3-6} \cmidrule(l){7-10}
    & Training data & None & 1 subject & 2 subjects & All & None & 1 subject & 2 subjects & All \\
    \midrule
     & Real data only & -- & 13.9 $\pm$ 0.59 & 12.4 $\pm$ 0.19 & \best{10.5\pm0.08} & -- & 45.8 $\pm$ 2.8 & 56.6 $\pm$ 2.6 & 74.8 $\pm$ 0.9 \\
    \midrule
    Matched & \projectname{} Garment & \best{14.8\pm0.05} & \best{13.1\pm0.56} & \best{12.0\pm0.30} & \best{10.5\pm0.04} & \best{32.3\pm1.4} & \best{54.9\pm2.2} & \best{64.0\pm0.4} & \best{77.7\pm0.9} \\
    \midrule
    Generic & \projectname{} Garment & 15.1 $\pm$ 0.17 & 13.3 $\pm$ 0.50 & 12.2 $\pm$ 0.19 & 10.6 $\pm$ 0.01 & 28.1 $\pm$ 2.8 & 53.9 $\pm$ 1.3 & 63.9 $\pm$ 1.3 & 76.7 $\pm$ 0.4 \\
    \bottomrule
  \end{tabular}
\end{table}

%% file: sections/conclusion/main.tex
\section{Conclusion}
\label{sec:conclusion}
We presented \projectname{}, a pipeline that generates garment-mounted virtual IMU data from text by physically simulating garments that are matched to the clothing and sensor layout of a target deployment, and that yields paired garment and body signals for every simulated motion.
In a large-scale evaluation on four real datasets and two tasks, \projectname{} Body + Garment data improved activity recognition over body-mounted virtual data when real data came from one subject, and \projectname{} Garment data lowered pose error from real loose-garment IMUs by 12.1\% relative to body-mounted virtual data when the garment was simulated on recorded motion.
Garment physics matters most when real data are scarce, realistic motion increases its benefit further, and for activity recognition, \projectname{} Garment and \projectname{} Body + Garment data remain the strongest virtual sources even when real data are plentiful.
\projectname{} thus turns virtual IMU data from a proxy for body-mounted sensors into a model of the smart garment itself.
It reduces the amount of real data that garment-mounted sensing requires, and it lays the groundwork for researchers who want to study smart-clothing sensing but have no recorded data of their own.

%% file: sections/appendix/main.tex
\section{Simulation Details}
\label{app:sim}
This appendix gives further details and the key settings behind Section~\ref{sec:system}.

\subsection{Garments}
\label{app:garments}
For every dataset, we generated a garment that matches what its participants wore (Figure~\ref{fig:garments}, Table~\ref{tab:garments}), from photographs of the dataset's garment or, where no usable photograph exists, from the published description.
Patterns are sized to the SMPL body with $\beta=0$, and tight sleeves keep at least 3\,cm of ease down to the wrist.
Fabric parameters are set per garment from the fabric type (Table~\ref{tab:sim_params}).

\input{figure/garments/garments}

\input{tables/garments}

\input{tables/sim_params}

\subsection{Sensor Mapping}
\label{app:sites}
Each real sensor is defined by a vertex of the SMPL body mesh and the garment it is sewn onto, and its virtual counterpart is the point of that garment closest to the body vertex in the start state.
For MIG, we place all 396 sensors from the dataset's sensor-position photographs.
The sensor positions of all four datasets are marked on their garments in Figure~\ref{fig:garments}, in a front and a back view, since many of them sit on the back and on the outer forearm.

\subsection{Pipeline Details}
\label{app:pipeline}
\paragraph{Sewing-pattern estimation.}
ChatGarment~\cite{chatgarment} is the LLaVA vision--language model~\cite{liu2023llava}, fine-tuned with LoRA~\cite{hu2022lora}.
Given the image $X_v$ and the text prompt $X_q$, the language model $f_\phi$ writes a response $Y_t = f_\phi(X_v, X_q)$, or $Y_t = f_\phi(X_q)$ when there is no image.
A small network $g_\Theta$ reads the continuous values off the model's internal state $H_t$ at the end of the response, $\mathbf n = g_\Theta(H_t) \in [0,1]^{76}$, with every value scaled to $[0,1]$.
The model was trained on 40{,}000 synthetic garments with known designs, with a text loss on the discrete choices and an $L_1$ loss on the continuous values.

\paragraph{Virtual sewing.}
Each vertex $i$ on one stitched edge is linked to the matching point on the other edge, which usually falls between two vertices $j$ and $k$ of that edge, at $(1-w)\,\mathbf x_j + w\,\mathbf x_k$ with $w\in[0,1]$.
Each solver step of the sewing stage finds
\begin{equation}
  \mathbf x^\ast = \arg\min_{\mathbf x}\; \Psi_{\mathrm{el}}(\mathbf x) + \sum_i \psi_{\mathrm{weak}}(g_i, \hat g, \bar\kappa_i) + \sum_{(i,j,k,w)} \frac{\kappa_s}{2}\,\big\|\mathbf x_i - (1-w)\,\mathbf x_j - w\,\mathbf x_k\big\|^2,
\end{equation}
where $\Psi_{\mathrm{el}}$ is the elastic energy of the fabric, the barriers $\psi_{\mathrm{weak}}$ keep the garment outside the body (Equation~\ref{eq:barrier}) and the last term pulls every stitch shut with stiffness $\kappa_s$.

\paragraph{Arm--torso clearance.}
Let $d_v(t)$ be the distance from an arm vertex $v$ to the torso in frame $t$, $\mathbf n_v$ the unit vector from $v$ toward the closest torso point and $d^\ast$ the target clearance.
Where the arm is closer to the torso than $d^\ast$, the arm vertex moves away from the torso by a fraction $\alpha$ of the missing clearance, and the closest torso vertex $u$ moves away from the arm by the remaining fraction,
\begin{equation}
  \Delta\mathbf x_v(t) = -\,\alpha\,\big[d^\ast - d_v(t)\big]_+\,\mathbf n_v, \qquad
  \Delta\mathbf x_u(t) = -\,(1-\alpha)\,\big[d^\ast - d_u(t)\big]_+\,\mathbf n_u ,
\end{equation}
where $\mathbf n_u$ points from $u$ toward the arm, $[\cdot]_+=\max(\cdot,0)$ and $\alpha=0.5$ shares the correction equally.

\paragraph{T-pose calibration.}
The virtual wearer holds a T-pose before the motion, and the garment settles on the raised arms.
The stored rotation $O_T$ is the average of $B(t)^\top S(t)$ at the end of that hold, where $S(t)$ is the orientation of the garment sensor and $B(t)$ that of the body sensor beneath it, and the reported orientation is $S(t)\,O_T^\top$.
Because raising the arms changes how a sleeve hangs, this simulated hold matters most for sensors on the sleeves.

\subsection{Simulation Parameters}
\label{app:collider}
\label{app:dt}
Table~\ref{tab:sim_params} lists the fabric of each garment and the settings shared by all simulations.
The collider settings follow from the body. With the default SMPL body shape, the upper arm of a generated walk sits about 22\,mm inside the torso, because SMPL's torso is fuller than that of the body the motion model was trained on. The collider therefore pushes arm and torso apart to at least 8\,mm clearance, without moving any sensor position.
The creases that linear blend skinning forms at flexed elbows, at the hip and at the crotch are rounded with a radius of 12--22\,mm at the elbow and 15--35\,mm at the hip, which changes the garment IMU only within the run-to-run noise of the solver.
Each motion is preceded by a 1.5\,s transition from the dressing pose into its first frame.
The solver uses a time step of 1/240\,s with output at 120\,Hz, a friction coefficient of 0.5, stiffness-proportional damping of 0.02\,s and a pin that holds the trousers at the waist like a belt.
It sets the stiffness of the barrier of each contact $i$ (Equation~\ref{eq:barrier}) as
\begin{equation}
  \bar\kappa_i = \frac{m_i}{g_i^2} + \frac{\mathbf w_i}{\|\mathbf w_i\|}\cdot\Big(H\,\frac{\mathbf w_i}{\|\mathbf w_i\|}\Big),
\end{equation}
where $m_i$ is the mass of the contacting vertices, $\mathbf w_i$ the contact direction distributed to them and $H$ the Hessian of the elastic energy~\cite{ppf_solver}.
The first term stops the fabric's momentum within the gap, and the second resists the elastic force that presses the fabric onto the body.

\section{Datasets}
\label{app:datasets}
Table~\ref{tab:datasets} summarizes the four datasets of Section~\ref{sec:eval:setup}, their activities and the amount of real and virtual data. Figure~\ref{fig:renders} shows examples of the virtual data, the simulated garments in motion.

\input{tables/datasets}
\input{figure/renders/renders}

\section{Evaluation Details}
\label{app:protocol}
\label{app:models}
We use three activity recognition models, DeepConvLSTM~\cite{deepconvlstm}, DeepConvLSTM with attention~\cite{murahari2018attention} and TinyHAR~\cite{zhou2022tinyhar}, and two pose estimation networks, an LSTM poser and TransPose~\cite{yi2021transpose}.
This section describes their inputs and architectures and how they are trained.

\paragraph{Activity recognition.}
Real and virtual signals are resampled to 20\,Hz and cut into 2\,s windows (40 samples) with 50\% overlap.
Every sensor contributes its accelerometer (LIP, FIP and GarMoCap) or its accelerometer and gyroscope (MIG) in its own frame.
The datasets' activity classes are used as published, and every generated clip is labeled with the class of its description.
DeepConvLSTM has four convolutional layers of 64 filters with kernel size 5, followed by a two-layer LSTM with 128 units and dropout of 0.5.
DeepConvLSTM with attention uses the same trunk and adds a temporal self-attention head over the LSTM outputs, whose result is added to the output of the last time step.
TinyHAR processes each channel with four convolutional layers of 20 filters (kernel size 5, strides 1, 2, 1 and 2, batch normalization), lets the channels interact through a transformer layer, fuses them with a fully connected layer and aggregates over time with an LSTM and temporal attention.
All three are trained with Adam at a learning rate of $10^{-3}$ and a batch size of 128, for 30 epochs on the real and virtual windows together.

\paragraph{Pose estimation.}
LIP and FIP sequences are used at 30\,Hz and GarMoCap sequences at 60\,Hz.
The input of every frame is the free acceleration and the orientation of each garment sensor relative to the root sensor, and the target is the SMPL joint rotations and joint positions.
Virtual orientations are calibrated as the datasets calibrate their real ones, with a simulated T-pose calibration on all three datasets (Section~\ref{sec:system:sensing}).
The LSTM poser has the shape of the network of Loose Inertial Poser~\cite{lip_cvpr2024}. A linear layer with 256 units, tanh activation and dropout of 0.2 feeds a two-layer LSTM with 256 units, which regresses the joint rotations and positions of every frame.
TransPose is used with its pose branch of three bidirectional LSTM stages (hidden sizes 256, 64 and 128). They estimate the leaf-joint positions, then all joint positions, then the joint rotations, with noise added to the intermediate positions during training as in the original work. It runs online with a window of 20 past and 5 future frames.
Both networks are trained on random 128-frame crops with a batch size of 64 and Adam at a learning rate of $10^{-3}$, reduced by a factor of 0.3 halfway. The loss is the error of the rotations plus ten times the error of the positions.
On LIP and FIP for both networks, and on GarMoCap for TransPose, pre-training runs for 2{,}000 or 5{,}000 iterations and fine-tuning on real data for 2{,}000 or 4{,}000 iterations at 0.1 or 0.3 times the learning rate. These three settings are tuned as hyperparameters per network and amount of real data; without real data, 2{,}000 iterations of pre-training are used throughout.
On GarMoCap, the LSTM poser is instead trained on 256-frame crops for 6{,}000 iterations and fine-tuned for 3{,}000 iterations at a learning rate of $3\times10^{-4}$.

\section{Per-Model Results}
\label{app:har}
\label{app:pose}
Tables~\ref{tab:har_permodel} and~\ref{tab:har_permodel_b} (recognition) and Tables~\ref{tab:pose_permodel} and~\ref{tab:pose_permodel_pos} (pose: joint angle error and MPJPE) report the results of each recognition model and each pose network behind Tables~\ref{tab:har_main} and~\ref{tab:pose_main}.

\input{tables/har_permodel}

\input{tables/pose_permodel}

%% file: figure/garments/garments.tex
\begin{figure*}[!tbp]
  \centering
  \includegraphics[width=\linewidth]{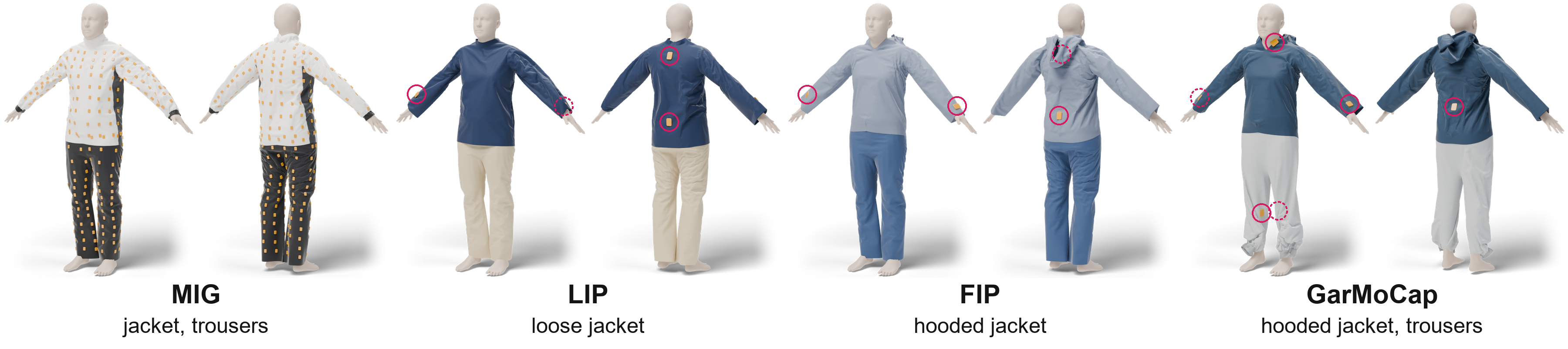}
  \caption{The garments matched to the four datasets, each sewn onto the body and settled in the A-pose, in a front and a back view. Orange boards mark the sensor positions of the matched dataset (all 396 for MIG). For LIP, FIP and GarMoCap, each position is ringed once: solid in the view where its board is visible, and dashed where the board is hidden in both views, such as on the underside of a sleeve or under the hood, in the view it faces most. MIG: sensor jacket and trousers. LIP: loose outdoor jacket. FIP: loose hooded jacket. GarMoCap: hooded jacket and cuffed trousers.}
  \label{fig:garments}
\end{figure*}

%% file: tables/garments.tex
\begin{table}[!tbp]
  \caption{Garments matched to the four datasets. Each garment is reconstructed from the dataset's garment and carries the dataset's sensor positions.}
  \label{tab:garments}
  \centering
  \footnotesize
  \begin{tabular}{@{}lll@{}}
    \toprule
    Matched dataset & Garment & Real sensors \\
    \midrule
    MIG~\cite{mig_dataset} & jacket + trousers & 396 sewn IMUs \\
    LIP~\cite{lip_cvpr2024} & loose jacket + trousers & 4 (cuffs, back, low back) \\
    FIP~\cite{fip_dataset} & loose hooded jacket +  trousers & 4 (forearms, back, waist) \\
    GarMoCap~\cite{garmocap_gid} & loose hooded jacket + trousers & 6 (cuffs, back, hem, ankles) \\
    \bottomrule
  \end{tabular}
\end{table}

%% file: tables/sim_params.tex
\begin{table}[!tbp]
  \caption{Simulation parameters. Top: fabric of the shell of each matched garment (areal density, bending stiffness $B$) and of the trousers. Bottom: solver and sensor settings shared by all garments.}
  \label{tab:sim_params}
  \centering
  \small
  \begin{tabular}{@{}llcc@{}}
    \toprule
    Garment & Shell fabric & Density (g/m$^2$) & $B$ ($10^{-6}$\,N\,m) \\
    \midrule
    MIG jacket & Polyester windbreaker taffeta & 90 & 4.7 \\
    LIP jacket & Light polyester woven & 112 & 4.6 \\
    FIP jacket & Nylon shell with polyester lining & 200 & 11 \\
    GarMoCap jacket & Polyester windbreaker taffeta & 90 & 4.7 \\
    Trousers (all) & Polyester--cotton twill & 230 & 24 \\
    \midrule
    \multicolumn{4}{@{}l}{Time step 1/240\,s, output 120\,Hz; friction coefficient 0.5; stiffness-proportional damping 0.02\,s} \\
    \multicolumn{4}{@{}l}{Settling 0.2\,s; 1.5\,s transition from the dressing pose; trousers pinned at the waist like a belt} \\
    \multicolumn{4}{@{}l}{Arm--torso clearance of the collider at least 8\,mm; elbow and hip crease fillets of 12--35\,mm radius} \\
    \multicolumn{4}{@{}l}{Rigid sensor patch of radius 2.5\,cm on garment and body} \\
    \bottomrule
  \end{tabular}
\end{table}

%% file: tables/datasets.tex
\begin{table}[!htbp]
  \caption{Datasets used in the evaluation and the virtual data generated for each. Real: recorded data with garment-mounted IMUs. Virtual: simulated motion on the garment matched to the dataset's clothing. Real and virtual amounts in minutes.}
  \label{tab:datasets}
  \centering
  \footnotesize
  \setlength{\tabcolsep}{4pt}
  \begin{tabular}{@{}lccllrr@{}}
    \toprule
    Dataset & Subjects & IMUs & Modalities & Activities & Real & Virtual \\
    \midrule
    MIG~\cite{mig_dataset} & 12 & 396 (6 used) & Acc., gyro. & \parbox[t]{4.5cm}{\raggedright Eating a cookie, eating pasta, folding, going downstairs, going upstairs, ironing, jogging, sitting, standing, toothbrushing, typing, vacuuming, walking, writing} & 432 & 89 \\[1.5ex]
    LIP~\cite{lip_cvpr2024} & 5 & 4 & Acc., orientation & \parbox[t]{4.5cm}{\raggedright Running, walking, jumping, boxing, ping-pong} & 118 & 12 \\
    FIP~\cite{fip_dataset} & 10 & 4 & Acc., orientation & \parbox[t]{4.5cm}{\raggedright Walking, running, boxing, jumping rope, chest expansion, lifting weights, dribbling, drumming, crossing hands, ping-pong} & 103 & 24 \\[1.5ex]
    GarMoCap~\cite{garmocap_gid} & 9 & 6 & Acc., orientation & \parbox[t]{4.5cm}{\raggedright Basketball, boxing, casual movement, jumping, jumping rope, ping-pong, rotating, running, tennis, walking} & 93 & 21 \\[1.5ex]
    \bottomrule
  \end{tabular}
\end{table}

%% file: figure/renders/renders.tex
\begin{figure}[!tbp]
  \centering
  \includegraphics[width=0.8\linewidth]{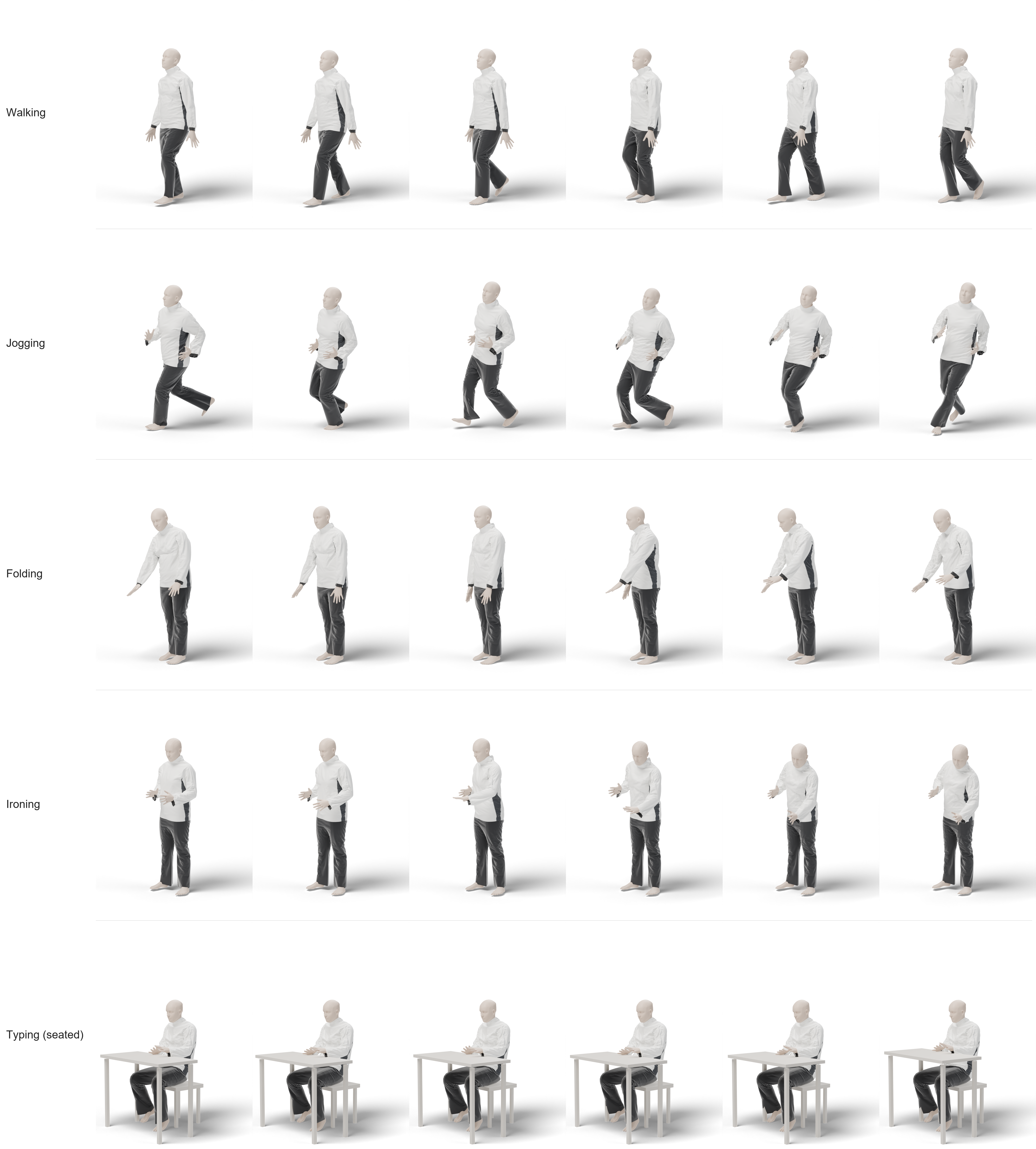}
  \caption{\projectname{} garments in motion. Each row shows frames of one generated activity with the physics-simulated jacket and trousers. }
  \label{fig:renders}
\end{figure}

%% file: tables/har_permodel.tex
\begin{table}[!tbp]
  \caption{Activity recognition performance of each model on the LIP and FIP datasets (macro-F1, \%): DeepConvLSTM (DCL), DeepConvLSTM with attention (DCL-Att) and TinyHAR, averaged over the test subjects. The best result in each block and column is highlighted.}
  \label{tab:har_permodel}
  \centering
  \scriptsize
  \setlength{\tabcolsep}{2.5pt}
  \begin{tabular}{@{}lcccccc@{}}
    \toprule
    & \multicolumn{3}{c}{LIP} & \multicolumn{3}{c}{FIP} \\
    \cmidrule(lr){2-4}\cmidrule(lr){5-7}
    Training data & DCL & TinyHAR & DCL-Att & DCL & TinyHAR & DCL-Att \\
    \midrule
    \multicolumn{7}{c}{Virtual data only} \\
    \midrule
    IMUGPT & 48.3 $\pm$ 1.8 & 59.5 $\pm$ 2.5 & 47.7 $\pm$ 6.4 & 15.2 $\pm$ 1.1 & 17.5 $\pm$ 0.9 & 15.8 $\pm$ 1.4 \\
    TailorNet & 66.9 $\pm$ 1.8 & 66.1 $\pm$ 2.7 & 62.5 $\pm$ 5.5 & 21.1 $\pm$ 1.9 & 20.5 $\pm$ 1.0 & 20.8 $\pm$ 0.5 \\
    \projectname{} Body & 65.8 $\pm$ 0.9 & 66.0 $\pm$ 3.6 & 65.3 $\pm$ 9.7 & 16.8 $\pm$ 2.6 & 20.5 $\pm$ 1.1 & 18.8 $\pm$ 1.6 \\
    \projectname{} Garment & 54.3 $\pm$ 3.9 & 64.4 $\pm$ 4.6 & 57.6 $\pm$ 4.3 & 20.6 $\pm$ 2.0 & 22.9 $\pm$ 1.1 & 20.7 $\pm$ 2.2 \\
    \projectname{} Body + Garment & \best{71.4\pm2.8} & \best{71.5\pm2.2} & \best{73.0\pm5.5} & \best{23.4\pm0.8} & \best{24.0\pm0.7} & \best{22.0\pm1.1} \\
    \midrule
    \multicolumn{7}{c}{1 real subject} \\
    \midrule
    Real data only & 61.3 $\pm$ 3.0 & 75.5 $\pm$ 5.8 & 63.9 $\pm$ 4.8 & 15.8 $\pm$ 2.3 & 21.2 $\pm$ 8.5 & 16.0 $\pm$ 2.7 \\
    IMUGPT & 58.0 $\pm$ 2.4 & 74.3 $\pm$ 6.5 & 66.8 $\pm$ 6.0 & 15.7 $\pm$ 2.5 & 20.7 $\pm$ 6.2 & 17.8 $\pm$ 1.6 \\
    TailorNet & 75.9 $\pm$ 4.9 & 76.2 $\pm$ 1.9 & 77.3 $\pm$ 1.5 & 22.8 $\pm$ 1.9 & 27.1 $\pm$ 2.9 & 21.9 $\pm$ 0.2 \\
    \projectname{} Body & 79.4 $\pm$ 8.0 & 78.7 $\pm$ 1.3 & 76.5 $\pm$ 6.5 & 22.8 $\pm$ 0.6 & 26.7 $\pm$ 5.4 & 20.4 $\pm$ 1.8 \\
    \projectname{} Garment & 76.3 $\pm$ 2.1 & \best{81.6\pm4.5} & 75.6 $\pm$ 5.0 & 24.2 $\pm$ 1.4 & \best{30.9\pm2.6} & 21.7 $\pm$ 2.1 \\
    \projectname{} Body + Garment & \best{80.0\pm3.6} & 80.1 $\pm$ 3.3 & \best{79.8\pm4.5} & \best{25.3\pm1.0} & 29.9 $\pm$ 3.0 & \best{22.5\pm1.0} \\
    \midrule
    \multicolumn{7}{c}{2 real subjects} \\
    \midrule
    Real data only & 76.9 $\pm$ 3.6 & 83.1 $\pm$ 2.6 & 80.2 $\pm$ 4.3 & 27.9 $\pm$ 4.1 & 32.7 $\pm$ 3.6 & 26.3 $\pm$ 5.0 \\
    IMUGPT & 75.4 $\pm$ 2.8 & 84.3 $\pm$ 2.6 & 75.8 $\pm$ 3.9 & 25.7 $\pm$ 3.2 & 30.6 $\pm$ 6.3 & 24.6 $\pm$ 1.1 \\
    TailorNet & 82.9 $\pm$ 1.4 & 83.3 $\pm$ 3.9 & 80.4 $\pm$ 2.6 & 32.1 $\pm$ 2.7 & 34.2 $\pm$ 4.3 & 31.4 $\pm$ 4.9 \\
    \projectname{} Body & \best{86.0\pm0.7} & 83.9 $\pm$ 2.8 & \best{84.3\pm1.9} & 33.4 $\pm$ 3.3 & 34.6 $\pm$ 2.8 & 31.3 $\pm$ 0.7 \\
    \projectname{} Garment & 84.6 $\pm$ 2.2 & \best{87.2\pm2.1} & 83.1 $\pm$ 5.4 & 33.2 $\pm$ 2.6 & \best{38.5\pm2.7} & 33.9 $\pm$ 2.1 \\
    \projectname{} Body + Garment & 84.2 $\pm$ 1.8 & 86.0 $\pm$ 0.8 & 84.2 $\pm$ 2.9 & \best{34.9\pm2.3} & 37.0 $\pm$ 2.3 & \best{34.6\pm1.9} \\
    \midrule
    \multicolumn{7}{c}{All real subjects} \\
    \midrule
    Real data only & 85.7 $\pm$ 1.2 & 89.4 $\pm$ 1.3 & 87.2 $\pm$ 0.3 & 56.8 $\pm$ 0.7 & 54.5 $\pm$ 0.7 & 55.9 $\pm$ 1.9 \\
    IMUGPT & 85.5 $\pm$ 1.5 & \best{90.7\pm1.5} & 85.8 $\pm$ 4.4 & 55.2 $\pm$ 1.1 & 53.7 $\pm$ 1.6 & 53.4 $\pm$ 1.4 \\
    TailorNet & 86.6 $\pm$ 1.8 & 90.5 $\pm$ 1.1 & 87.6 $\pm$ 1.4 & 57.8 $\pm$ 0.6 & 53.4 $\pm$ 0.6 & 56.6 $\pm$ 1.2 \\
    \projectname{} Body & \best{87.6\pm0.7} & 90.6 $\pm$ 0.6 & 86.7 $\pm$ 2.7 & 56.5 $\pm$ 2.1 & 51.5 $\pm$ 1.5 & 54.0 $\pm$ 2.0 \\
    \projectname{} Garment & 87.6 $\pm$ 1.6 & 89.8 $\pm$ 1.0 & \best{88.0\pm0.7} & 56.3 $\pm$ 2.7 & \best{55.7\pm0.7} & \best{57.7\pm0.5} \\
    \projectname{} Body + Garment & 87.1 $\pm$ 0.3 & 89.6 $\pm$ 1.5 & 87.9 $\pm$ 2.8 & \best{59.0\pm2.1} & 55.6 $\pm$ 1.8 & 56.9 $\pm$ 0.6 \\
    \bottomrule
  \end{tabular}
\end{table}

\begin{table}[!tbp]
  \caption{Activity recognition performance of each model on the GarMoCap and MIG datasets (macro-F1, \%): DeepConvLSTM (DCL), DeepConvLSTM with attention (DCL-Att) and TinyHAR, averaged over the test subjects. The best result in each block and column is highlighted.}
  \label{tab:har_permodel_b}
  \centering
  \scriptsize
  \setlength{\tabcolsep}{2.5pt}
  \begin{tabular}{@{}lcccccc@{}}
    \toprule
    & \multicolumn{3}{c}{GarMoCap} & \multicolumn{3}{c}{MIG} \\
    \cmidrule(lr){2-4}\cmidrule(lr){5-7}
    Training data & DCL & TinyHAR & DCL-Att & DCL & TinyHAR & DCL-Att \\
    \midrule
    \multicolumn{7}{c}{Virtual data only} \\
    \midrule
    IMUGPT & 20.9 $\pm$ 1.2 & 30.7 $\pm$ 3.7 & 21.3 $\pm$ 0.8 & 5.0 $\pm$ 0.4 & 11.0 $\pm$ 0.6 & 5.7 $\pm$ 1.5 \\
    TailorNet & 25.1 $\pm$ 1.4 & \best{33.8\pm1.9} & \best{27.2\pm1.9} & 20.2 $\pm$ 2.7 & 39.9 $\pm$ 3.7 & 22.4 $\pm$ 4.9 \\
    \projectname{} Body & \best{28.0\pm1.4} & 33.0 $\pm$ 3.6 & 26.3 $\pm$ 3.1 & 35.0 $\pm$ 0.9 & 44.9 $\pm$ 0.4 & 33.3 $\pm$ 1.5 \\
    \projectname{} Garment & 24.9 $\pm$ 2.9 & 33.5 $\pm$ 0.5 & 21.5 $\pm$ 1.6 & 40.9 $\pm$ 0.8 & 47.1 $\pm$ 3.1 & 42.6 $\pm$ 0.9 \\
    \projectname{} Body + Garment & 25.8 $\pm$ 4.8 & 32.0 $\pm$ 2.3 & 24.8 $\pm$ 1.6 & \best{41.0\pm1.3} & \best{52.4\pm0.8} & \best{43.5\pm1.7} \\
    \midrule
    \multicolumn{7}{c}{1 real subject} \\
    \midrule
    Real data only & 40.3 $\pm$ 2.6 & \best{54.3\pm4.3} & 42.8 $\pm$ 2.4 & 34.1 $\pm$ 0.7 & 48.0 $\pm$ 3.8 & 37.3 $\pm$ 4.6 \\
    IMUGPT & 37.6 $\pm$ 1.8 & 49.2 $\pm$ 3.4 & 36.8 $\pm$ 1.9 & 27.5 $\pm$ 3.1 & 47.0 $\pm$ 4.1 & 28.3 $\pm$ 4.4 \\
    TailorNet & 45.8 $\pm$ 1.1 & 52.4 $\pm$ 1.7 & 44.4 $\pm$ 3.6 & 38.8 $\pm$ 2.3 & 51.9 $\pm$ 3.5 & 40.4 $\pm$ 3.2 \\
    \projectname{} Body & 43.5 $\pm$ 1.2 & 53.6 $\pm$ 1.5 & 44.7 $\pm$ 0.6 & 45.4 $\pm$ 0.8 & 55.3 $\pm$ 3.5 & 45.4 $\pm$ 2.3 \\
    \projectname{} Garment & 46.1 $\pm$ 1.2 & 52.6 $\pm$ 0.3 & 45.2 $\pm$ 3.9 & \best{45.9\pm4.5} & 55.4 $\pm$ 4.3 & \best{48.2\pm1.8} \\
    \projectname{} Body + Garment & \best{48.5\pm2.9} & 53.2 $\pm$ 1.3 & \best{45.6\pm3.6} & 45.2 $\pm$ 6.5 & \best{56.1\pm5.2} & 47.8 $\pm$ 6.6 \\
    \midrule
    \multicolumn{7}{c}{2 real subjects} \\
    \midrule
    Real data only & 53.8 $\pm$ 4.7 & 61.6 $\pm$ 2.0 & 54.4 $\pm$ 2.5 & 44.9 $\pm$ 4.6 & 59.6 $\pm$ 1.5 & 45.1 $\pm$ 1.0 \\
    IMUGPT & 51.7 $\pm$ 1.2 & 60.4 $\pm$ 1.9 & 51.3 $\pm$ 2.7 & 40.4 $\pm$ 0.8 & 57.9 $\pm$ 2.8 & 42.3 $\pm$ 4.2 \\
    TailorNet & 53.3 $\pm$ 0.7 & 61.2 $\pm$ 3.9 & 56.0 $\pm$ 3.6 & 48.9 $\pm$ 1.5 & 58.4 $\pm$ 1.9 & 45.4 $\pm$ 5.2 \\
    \projectname{} Body & 53.9 $\pm$ 3.4 & 60.6 $\pm$ 1.9 & 55.9 $\pm$ 2.4 & 47.9 $\pm$ 1.3 & 60.5 $\pm$ 2.4 & 49.4 $\pm$ 2.5 \\
    \projectname{} Garment & \best{55.4\pm1.9} & 64.1 $\pm$ 1.8 & 54.8 $\pm$ 4.5 & \best{53.7\pm0.9} & \best{63.3\pm1.6} & \best{54.8\pm2.1} \\
    \projectname{} Body + Garment & 53.3 $\pm$ 4.6 & \best{64.3\pm1.5} & \best{58.3\pm0.6} & 52.5 $\pm$ 2.6 & 62.6 $\pm$ 1.5 & 54.2 $\pm$ 1.1 \\
    \midrule
    \multicolumn{7}{c}{All real subjects} \\
    \midrule
    Real data only & 75.3 $\pm$ 2.1 & 75.3 $\pm$ 0.7 & 73.7 $\pm$ 1.1 & 65.8 $\pm$ 1.8 & 69.2 $\pm$ 0.4 & 66.9 $\pm$ 1.0 \\
    IMUGPT & 73.2 $\pm$ 1.9 & 76.4 $\pm$ 0.8 & 73.8 $\pm$ 0.5 & 64.2 $\pm$ 1.0 & 69.7 $\pm$ 1.6 & 62.4 $\pm$ 2.5 \\
    TailorNet & 74.1 $\pm$ 2.2 & 76.0 $\pm$ 1.8 & 74.6 $\pm$ 0.7 & 65.7 $\pm$ 1.1 & 71.3 $\pm$ 0.6 & 67.2 $\pm$ 2.5 \\
    \projectname{} Body & 73.6 $\pm$ 1.0 & 75.2 $\pm$ 1.1 & 73.3 $\pm$ 2.3 & 66.6 $\pm$ 3.5 & 71.2 $\pm$ 0.7 & 65.5 $\pm$ 0.4 \\
    \projectname{} Garment & \best{76.3\pm3.2} & \best{77.8\pm0.3} & \best{76.2\pm2.5} & \best{69.3\pm3.4} & 73.1 $\pm$ 1.4 & \best{70.5\pm0.9} \\
    \projectname{} Body + Garment & 76.1 $\pm$ 2.5 & 75.9 $\pm$ 1.8 & 75.2 $\pm$ 1.3 & 67.8 $\pm$ 2.0 & \best{73.6\pm1.4} & 70.3 $\pm$ 2.6 \\
    \bottomrule
  \end{tabular}
\end{table}

%% file: tables/pose_permodel.tex
\begin{table}[!tbp]
  \caption{Pose estimation performance of each network: mean angular error over 15 major joints (degrees), lower is better. The best result in each block and column is highlighted.}
  \label{tab:pose_permodel}
  \centering
  \footnotesize
  \setlength{\tabcolsep}{3pt}
  \begin{tabular}{@{}lcccccc@{}}
    \toprule
    & \multicolumn{2}{c}{LIP} & \multicolumn{2}{c}{FIP} & \multicolumn{2}{c}{GarMoCap} \\
    \cmidrule(lr){2-3}\cmidrule(lr){4-5}\cmidrule(lr){6-7}
    Training data & LSTM & TransPose & LSTM & TransPose & LSTM & TransPose \\
    \midrule
    \multicolumn{7}{c}{Virtual data only} \\
    \midrule
    IMUGPT & 33.5 $\pm$ 1.24 & 29.2 $\pm$ 1.07 & 28.3 $\pm$ 0.38 & 25.2 $\pm$ 0.49 & 24.2 $\pm$ 0.44 & 21.5 $\pm$ 0.18 \\
    TailorNet & 29.8 $\pm$ 0.38 & 28.7 $\pm$ 0.16 & 25.4 $\pm$ 0.50 & 25.6 $\pm$ 0.86 & 24.1 $\pm$ 0.77 & 22.7 $\pm$ 0.24 \\
    \projectname{} Body & 28.7 $\pm$ 0.57 & 28.5 $\pm$ 0.33 & 26.2 $\pm$ 0.53 & 25.5 $\pm$ 0.16 & 23.7 $\pm$ 0.62 & 23.2 $\pm$ 0.30 \\
    \projectname{} Garment & 28.8 $\pm$ 0.39 & 28.4 $\pm$ 0.59 & \best{23.1\pm0.68} & \best{24.2\pm0.25} & 21.9 $\pm$ 0.29 & \best{20.3\pm0.22} \\
    \projectname{} Body + Garment & \best{28.3\pm0.32} & \best{28.3\pm0.31} & 24.0 $\pm$ 0.48 & 24.7 $\pm$ 0.45 & \best{21.6\pm0.09} & 20.6 $\pm$ 0.11 \\
    \midrule
    \multicolumn{7}{c}{1 real subject} \\
    \midrule
    Real data only & 23.3 $\pm$ 1.17 & 22.5 $\pm$ 1.40 & 15.2 $\pm$ 1.11 & 14.9 $\pm$ 0.97 & 14.4 $\pm$ 0.61 & 13.5 $\pm$ 0.60 \\
    IMUGPT & 23.1 $\pm$ 1.20 & \best{22.3\pm1.23} & 15.5 $\pm$ 1.07 & 15.0 $\pm$ 0.94 & 14.4 $\pm$ 0.40 & 13.1 $\pm$ 0.53 \\
    TailorNet & \best{22.9\pm1.25} & 22.4 $\pm$ 1.09 & 15.4 $\pm$ 1.18 & 14.8 $\pm$ 0.71 & 14.2 $\pm$ 0.48 & 13.2 $\pm$ 0.55 \\
    \projectname{} Body & 23.0 $\pm$ 1.18 & 22.3 $\pm$ 1.09 & 15.6 $\pm$ 1.47 & 15.1 $\pm$ 1.12 & 14.3 $\pm$ 0.49 & 13.2 $\pm$ 0.46 \\
    \projectname{} Garment & 23.3 $\pm$ 1.33 & 22.4 $\pm$ 1.19 & \best{14.8\pm0.64} & \best{14.6\pm0.73} & 13.8 $\pm$ 0.58 & 12.9 $\pm$ 0.50 \\
    \projectname{} Body + Garment & 23.1 $\pm$ 1.36 & 22.5 $\pm$ 1.29 & 15.1 $\pm$ 0.89 & 14.8 $\pm$ 0.71 & \best{13.6\pm0.56} & \best{12.9\pm0.40} \\
    \midrule
    \multicolumn{7}{c}{2 real subjects} \\
    \midrule
    Real data only & 21.0 $\pm$ 0.48 & 20.3 $\pm$ 0.66 & 14.0 $\pm$ 0.38 & 13.7 $\pm$ 0.15 & 12.7 $\pm$ 0.11 & 12.1 $\pm$ 0.27 \\
    IMUGPT & 20.9 $\pm$ 0.35 & \best{20.0\pm0.41} & 14.1 $\pm$ 0.40 & 13.7 $\pm$ 0.22 & 12.6 $\pm$ 0.13 & 11.9 $\pm$ 0.24 \\
    TailorNet & 20.9 $\pm$ 0.44 & 20.1 $\pm$ 0.44 & 14.0 $\pm$ 0.43 & 13.7 $\pm$ 0.34 & 12.7 $\pm$ 0.23 & 11.8 $\pm$ 0.15 \\
    \projectname{} Body & 21.0 $\pm$ 0.45 & 20.3 $\pm$ 0.33 & 13.9 $\pm$ 0.33 & 13.7 $\pm$ 0.28 & 12.6 $\pm$ 0.15 & 11.8 $\pm$ 0.19 \\
    \projectname{} Garment & \best{20.8\pm0.46} & 20.1 $\pm$ 0.30 & \best{13.9\pm0.26} & \best{13.5\pm0.30} & 12.7 $\pm$ 0.26 & 11.8 $\pm$ 0.30 \\
    \projectname{} Body + Garment & 20.9 $\pm$ 0.37 & 20.2 $\pm$ 0.39 & 13.9 $\pm$ 0.35 & 13.5 $\pm$ 0.36 & \best{12.4\pm0.09} & \best{11.6\pm0.23} \\
    \midrule
    \multicolumn{7}{c}{All real subjects} \\
    \midrule
    Real data only & 19.4 $\pm$ 0.11 & 18.8 $\pm$ 0.07 & 12.6 $\pm$ 0.09 & 12.3 $\pm$ 0.02 & 11.0 $\pm$ 0.13 & 10.1 $\pm$ 0.04 \\
    IMUGPT & 19.3 $\pm$ 0.03 & 18.8 $\pm$ 0.21 & 12.6 $\pm$ 0.12 & 12.2 $\pm$ 0.04 & 11.1 $\pm$ 0.20 & 10.3 $\pm$ 0.08 \\
    TailorNet & 19.5 $\pm$ 0.07 & 18.9 $\pm$ 0.09 & 12.5 $\pm$ 0.11 & 12.2 $\pm$ 0.03 & 11.2 $\pm$ 0.04 & 10.3 $\pm$ 0.06 \\
    \projectname{} Body & \best{19.3\pm0.09} & 18.8 $\pm$ 0.10 & 12.6 $\pm$ 0.08 & 12.1 $\pm$ 0.06 & 10.9 $\pm$ 0.06 & 10.3 $\pm$ 0.03 \\
    \projectname{} Garment & 19.5 $\pm$ 0.16 & \best{18.7\pm0.08} & 12.4 $\pm$ 0.06 & \best{12.0\pm0.07} & 11.0 $\pm$ 0.08 & 10.2 $\pm$ 0.07 \\
    \projectname{} Body + Garment & 19.4 $\pm$ 0.04 & 18.8 $\pm$ 0.02 & \best{12.4\pm0.12} & 12.1 $\pm$ 0.06 & \best{10.8\pm0.09} & \best{10.1\pm0.14} \\
    \bottomrule
  \end{tabular}
\end{table}

\begin{table}[!tbp]
  \caption{Pose estimation performance of each network: mean joint position error (MPJPE, cm), lower is better. The best result in each block and column is highlighted.}
  \label{tab:pose_permodel_pos}
  \centering
  \footnotesize
  \setlength{\tabcolsep}{3pt}
  \begin{tabular}{@{}lcccccc@{}}
    \toprule
    & \multicolumn{2}{c}{LIP} & \multicolumn{2}{c}{FIP} & \multicolumn{2}{c}{GarMoCap} \\
    \cmidrule(lr){2-3}\cmidrule(lr){4-5}\cmidrule(lr){6-7}
    Training data & LSTM & TransPose & LSTM & TransPose & LSTM & TransPose \\
    \midrule
    \multicolumn{7}{c}{Virtual data only} \\
    \midrule
    IMUGPT & 12.2 $\pm$ 0.61 & 10.4 $\pm$ 0.48 & 9.1 $\pm$ 0.08 & 8.1 $\pm$ 0.23 & 11.6 $\pm$ 0.17 & 10.0 $\pm$ 0.04 \\
    TailorNet & 10.7 $\pm$ 0.18 & 10.3 $\pm$ 0.12 & 8.6 $\pm$ 0.11 & 8.6 $\pm$ 0.25 & 10.7 $\pm$ 0.40 & 10.3 $\pm$ 0.09 \\
    \projectname{} Body & 10.4 $\pm$ 0.29 & 10.3 $\pm$ 0.16 & 8.9 $\pm$ 0.21 & 8.6 $\pm$ 0.19 & 10.4 $\pm$ 0.15 & 10.2 $\pm$ 0.33 \\
    \projectname{} Garment & \best{9.7\pm0.11} & \best{9.6\pm0.10} & \best{7.9\pm0.38} & \best{8.1\pm0.13} & 9.9 $\pm$ 0.09 & \best{9.2\pm0.26} \\
    \projectname{} Body + Garment & 9.9 $\pm$ 0.04 & 9.9 $\pm$ 0.09 & 8.2 $\pm$ 0.09 & 8.3 $\pm$ 0.20 & \best{9.7\pm0.04} & 9.2 $\pm$ 0.17 \\
    \midrule
    \multicolumn{7}{c}{1 real subject} \\
    \midrule
    Real data only & 7.9 $\pm$ 0.35 & 7.6 $\pm$ 0.39 & 4.9 $\pm$ 0.47 & 4.7 $\pm$ 0.39 & 7.1 $\pm$ 0.33 & 6.6 $\pm$ 0.34 \\
    IMUGPT & 7.9 $\pm$ 0.38 & \best{7.5\pm0.37} & 5.1 $\pm$ 0.47 & 4.8 $\pm$ 0.41 & 7.0 $\pm$ 0.26 & 6.4 $\pm$ 0.24 \\
    TailorNet & \best{7.8\pm0.37} & 7.5 $\pm$ 0.38 & 5.0 $\pm$ 0.51 & 4.7 $\pm$ 0.30 & 6.9 $\pm$ 0.34 & 6.4 $\pm$ 0.38 \\
    \projectname{} Body & 7.9 $\pm$ 0.39 & 7.6 $\pm$ 0.37 & 5.1 $\pm$ 0.63 & 4.9 $\pm$ 0.52 & 7.0 $\pm$ 0.35 & 6.4 $\pm$ 0.34 \\
    \projectname{} Garment & 8.0 $\pm$ 0.43 & 7.5 $\pm$ 0.32 & \best{4.7\pm0.27} & \best{4.6\pm0.32} & 6.8 $\pm$ 0.31 & 6.3 $\pm$ 0.31 \\
    \projectname{} Body + Garment & 7.9 $\pm$ 0.48 & 7.6 $\pm$ 0.44 & 4.8 $\pm$ 0.42 & 4.7 $\pm$ 0.32 & \best{6.7\pm0.34} & \best{6.2\pm0.26} \\
    \midrule
    \multicolumn{7}{c}{2 real subjects} \\
    \midrule
    Real data only & \best{7.2\pm0.07} & 6.8 $\pm$ 0.16 & 4.3 $\pm$ 0.15 & 4.2 $\pm$ 0.09 & 6.2 $\pm$ 0.05 & 5.8 $\pm$ 0.06 \\
    IMUGPT & 7.2 $\pm$ 0.02 & 6.9 $\pm$ 0.06 & 4.4 $\pm$ 0.23 & 4.3 $\pm$ 0.09 & 6.1 $\pm$ 0.05 & 5.6 $\pm$ 0.04 \\
    TailorNet & 7.2 $\pm$ 0.12 & 6.8 $\pm$ 0.10 & 4.4 $\pm$ 0.13 & 4.3 $\pm$ 0.13 & 6.2 $\pm$ 0.16 & 5.6 $\pm$ 0.08 \\
    \projectname{} Body & 7.2 $\pm$ 0.06 & 6.9 $\pm$ 0.03 & 4.3 $\pm$ 0.19 & 4.3 $\pm$ 0.09 & 6.2 $\pm$ 0.06 & 5.6 $\pm$ 0.08 \\
    \projectname{} Garment & 7.2 $\pm$ 0.12 & \best{6.8\pm0.07} & 4.3 $\pm$ 0.09 & 4.2 $\pm$ 0.11 & 6.1 $\pm$ 0.18 & 5.6 $\pm$ 0.05 \\
    \projectname{} Body + Garment & 7.2 $\pm$ 0.02 & 6.9 $\pm$ 0.08 & \best{4.3\pm0.16} & \best{4.2\pm0.14} & \best{6.0\pm0.13} & \best{5.5\pm0.09} \\
    \midrule
    \multicolumn{7}{c}{All real subjects} \\
    \midrule
    Real data only & \best{6.6\pm0.03} & 6.4 $\pm$ 0.05 & 3.8 $\pm$ 0.03 & 3.7 $\pm$ 0.01 & 5.3 $\pm$ 0.03 & 4.9 $\pm$ 0.01 \\
    IMUGPT & 6.7 $\pm$ 0.05 & 6.4 $\pm$ 0.07 & 3.9 $\pm$ 0.04 & 3.7 $\pm$ 0.01 & 5.3 $\pm$ 0.08 & 5.0 $\pm$ 0.01 \\
    TailorNet & 6.7 $\pm$ 0.02 & 6.4 $\pm$ 0.04 & 3.9 $\pm$ 0.04 & 3.7 $\pm$ 0.02 & 5.4 $\pm$ 0.04 & 5.0 $\pm$ 0.01 \\
    \projectname{} Body & 6.6 $\pm$ 0.03 & 6.4 $\pm$ 0.01 & 3.9 $\pm$ 0.01 & 3.7 $\pm$ 0.02 & 5.3 $\pm$ 0.04 & 5.0 $\pm$ 0.04 \\
    \projectname{} Garment & 6.7 $\pm$ 0.08 & \best{6.4\pm0.06} & 3.8 $\pm$ 0.02 & \best{3.7\pm0.03} & 5.3 $\pm$ 0.06 & 4.9 $\pm$ 0.01 \\
    \projectname{} Body + Garment & 6.7 $\pm$ 0.03 & 6.4 $\pm$ 0.04 & \best{3.8\pm0.03} & 3.7 $\pm$ 0.03 & \best{5.2\pm0.05} & \best{4.9\pm0.02} \\
    \bottomrule
  \end{tabular}
\end{table}